\documentclass[11pt]{article}

\usepackage[preprint]{acl}

\usepackage{times}
\usepackage{latexsym}

\usepackage[T1]{fontenc}

\usepackage[utf8]{inputenc}

\usepackage{microtype}

\usepackage{inconsolata}

\usepackage{graphicx}

\usepackage{comment}
\usepackage{booktabs}
\usepackage{multirow}
\usepackage{microtype}
\usepackage{hyperref}
\usepackage{url}
\usepackage{graphicx}
\usepackage{amsmath}
\usepackage{tabularx}
\usepackage{adjustbox}

\usepackage{lineno}
\usepackage{subcaption}
\usepackage{color-edits}
\addauthor[Mike]{mf}{purple}
\addauthor[Steven]{sf}{blue}
\addauthor[Linda]{lz}{orange}

\newcommand\blfootnote[1]{%
  \begingroup
  \renewcommand\thefootnote{}\footnote{#1}%
  \addtocounter{footnote}{-1}%
  \endgroup
}

\title{Bringing Up a Bilingual BabyLM: Investigating Multilingual Language Acquisition Using Small-Scale Models}

\author{Linda Zeng, Steven Y. Feng, Michael C. Frank \\
Stanford University\\
\texttt{\{lindazeng,syfeng,mcfrank\}@stanford.edu} \\
}

\usepackage{color-edits}
\addauthor[Mike]{mf}{purple}
\addauthor[Steven]{sf}{blue}
\addauthor[Linda]{lz}{orange}

\begin{document}

\maketitle

\begin{abstract}
Multilingualism is incredibly common around the world, leading to many important theoretical and practical questions, including whether multilingual acquisition leads to delays in learning and whether there are better and worse ways to structure multilingual input. 
Many correlational studies address these questions, but it is surprisingly difficult to get definitive answers because children cannot be randomly assigned to be multilingual and data are typically not matched between languages. 
We create matched mono- and bilingual datasets using synthetic data and machine translation to simulate a variety of controlled exposure conditions, pairing English with Spanish, Chinese, and Dutch input. 
We train GPT-2 models on these datasets and evaluate their performance on perplexity, grammaticality, and semantic knowledge.
Across model scales, measures, and languages, bilingual models perform similarly to monolingual models in the first language, but show strong performance in the second as well. Control experiments show that our findings hold on real-world language data and even on shuffled second-language input. These results suggest that there are no strong differences between different bilingual exposure regimes, and bilingual input poses no in-principle challenges for agnostic statistical learners.
\blfootnote{\fontsize{8.5}{9}\selectfont Code and data: \href{https://github.com/lindazeng979/bilingual-babyLM}{github.com/lindazeng979/bilingual-babyLM}}
\end{abstract}

\section{Introduction}

\begin{figure}[t]
\begin{center}
\includegraphics[width=\linewidth]{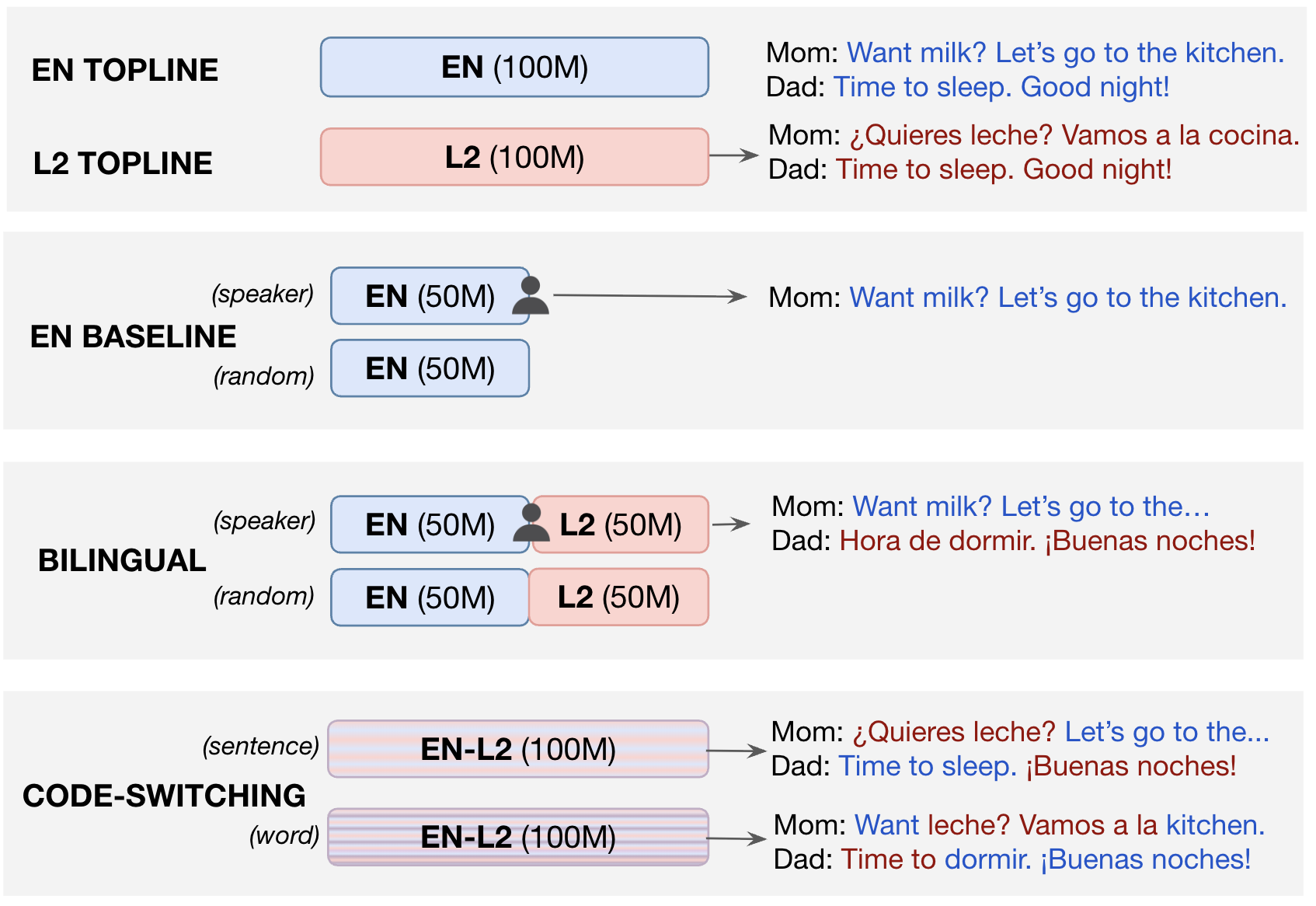}
\caption{Overview of matched language exposure conditions. Bilingual models are exposed to English (EN) and a second language (L2) simultaneously, rather than through sequential or curriculum-based training. In the ``speaker`` conditions, each speaker consistently uses one language (one speaker–one language policy), whereas in random conditions, either speaker may use either language, assigned randomly at the dialogue level. Examples are shown with Spanish as L2.}\label{fig:intro}
\end{center}
\end{figure}



A child growing up in a multilingual household may hear one language from a parent, another from a sibling, and a mixture of both during everyday conversation. To an outside observer, this linguistic environment may appear chaotic. Yet millions of children around the world successfully acquire multiple languages under these conditions. Multilingualism is a global norm: languages are used in fluid, dynamic ways that vary across speakers, contexts, and communities. 

Despite the prevalence of multilingual environments, the impacts of multilingual exposure have been hotly debated. Historically, multilingualism was considered negative (often on the basis of studies that confounded language exposure with other population risk factors), and concerns persist that simultaneous exposure to multiple languages may lead to developmental delays, language confusion, or other learning difficulties \citep{byers2013bilingualism,hoff2012dual}. These concerns are often discussed alongside prescriptive strategies such as the ``one-speaker–one-language'' approach \citep{ronjat1913developpement}, which aims to minimize potential confusion by strictly separating languages across caregivers. Similarly, code-switching, or alternating between languages, was historically discouraged, although research suggests that it is systematic, responsive to parental input, and does not necessarily impair language development \citep{comeau2003modeling,pearson2008raising,byers2017bilingual,kaushanskaya2023combining,libersky2024effects}.
 Meanwhile, some newer research has instead suggested potential cognitive benefits associated with multilingualism, including advantages in executive function, theory of mind, memory, and multitasking \citep{kovacs2009early, kovacs2009cognitive, goetz2003effects, brito2012influence, bialystok2012bilingualism, poulin2011effects,kaushanskaya2009bilingual,carlson2008bilingual}, though not all of this research has been replicable in subsequent investigations \citep{dick2019no,lehtonen2018bilingualism}.



Similar to debates over human multilingualism, the performance impacts of multilingual training on language models has been broadly debated. On one hand, multilingual models demonstrate strong cross-lingual transfer capabilities \citep{wu2019beto, wu-dredze-2020-languages,fujinuma-etal-2022-match,philippy-etal-2023-towards, 10.1145/3727339}. On the other hand, prior work has documented the ``curse of multilinguality,'' where training on many languages can degrade performance on high-resource languages due to capacity limitations or negative transfer \citep{ruder2017overview, wang-etal-2020-negative, conneau-etal-2020-unsupervised, pyysalo-etal-2021-wikibert, chang-etal-2024-multilinguality, papadimitriou-etal-2023-multilingual}. 

These findings raise questions that parallel those in developmental psychology: Does learning from multiple languages introduce interference, and does the structure of language exposure matter? More fundamentally, are these effects specific to multilingualism, or do they reflect the broader challenge of learning from two distinct input distributions? 
Ultimately, significant uncertainty remains regarding multilingual acquisition and the environments that best support it. 

Existing evidence comes from correlational studies of natural language environments, where experimental control is inherently limited: children cannot be randomly assigned to language exposure conditions, and it is impossible to control the quantity, content, and social context of linguistic input across individuals and languages \cite{de2019not}. Similarly, multilingual language model studies compare training conditions that differ in multiple dimensions, including data quality, semantic content, domain, discourse structure, and effective data quantity \citep{chang-etal-2024-multilinguality, constantinescu-etal-2025-investigating, jumelet2025babybabellm}. Consequently, it becomes difficult to isolate whether observed differences arise from multilingual exposure itself or properties of the training data.

Matched multilingual datasets provide a way forward to address this issue (Figure \ref{fig:intro}). Matched data can be used to hold the underlying linguistic content and total training budget fixed while systematically varying the languages and structures of exposure, isolating effects of language composition. We pursue this approach here, inspired by the  BabyLM paradigm, which attempts to model language learning under developmentally plausible data constraints (i.e., using 10M or 100M words of training data) \citep{warstadt-etal-2023-findings}. A variety of recent work has trained small language models (SLMs) on carefully constructed datasets to perform controlled simulations of language acquisition \citep{feng-etal-2024-child, kallini-etal-2024-mission, misra-mahowald-2024-language,hu2025production}. Building on this approach and prior work on multilingual training \citep{constantinescu-etal-2025-investigating, binyamin2026learningchilddirectedspeechtwolanguage}, we create synthetic data to simulate a variety of multilingual exposure conditions in English, Spanish, Chinese, and Dutch and measure their effects on models' learning outcomes. 

By studying language models as controlled statistical learners, we aim to track parallels and differences between multilingual learning in children and models. Language models here serve as distributional baselines for investigating which aspects of linguistic behavior emerge from statistical properties of the input alone \citep{trott2026large}. Rather than treating models as complete simulations of human learners, we use them to isolate how different multilingual exposure structures influence learning in the absence of many human-specific developmental mechanisms.
We investigate three research questions: \textbf{RQ1.} Does multilingual training induce interference or confer advantages relative to monolingual training? \textbf{RQ2.} How do different exposure structures -- such as speaker-separated input or code-switching -- affect learning outcomes? \textbf{RQ3.} How do these effects vary with model size, total data scale, and language proportion?

\section{Background and Related Work}
\label{gen_inst}

Current consensus suggests that multilingual exposure does not inherently impair language development or cause confusion between languages \citep{guiberson2013bilingual}. Within the first few months of life, infants can distinguish languages based on rhythmic and phonological cues \citep{mehler1988precursor, bosch1997native, nazzi2000language, byers2010roots}. Strict language separation strategies, such as ``one speaker, one language,'' appear neither harmful nor necessary, as speaker identity is only a weak cue for language differentiation \citep{de2007parental, potter2025infants}. Moreover, theoretical accounts propose that the demands of code-switching may shape executive function, although the nature and extent of these effects remain debated \citep{bialystok2017bilingual, arredondo2026missing}.

Multilingual training has shown both benefits, including cross-lingual transfer \citep{wu2019beto, wu-dredze-2020-languages, fujinuma-etal-2022-match, philippy-etal-2023-towards, 10.1145/3727339}, and drawbacks, including the ``curse of multilinguality,'' where additional languages can degrade performance due to capacity limitations or negative transfer \citep{ruder2017overview, wang-etal-2020-negative, conneau-etal-2020-unsupervised, pyysalo-etal-2021-wikibert, chang-etal-2024-multilinguality, papadimitriou-etal-2023-multilingual}. Practical challenges, including differences in tokenization  \citep{haslett-2025-tokenization, chang-etal-2024-multilinguality, wang2024tokenization, abagyan-etal-2026-one} and cross-lingual evaluation \citep{yang2026apples}, further complicate these comparisons. Here, our interest is not in building higher-performing multilingual models {\it per se} but rather in understanding and quantifying the impacts of varying multilingual exposures. 

While neural network models have long been used to explore multilingual learning \citep{li2004early}, SLMs in particular provide a tractable framework for investigating questions about the impacts of language training data under controlled input conditions \citep{huebner-etal-2021-babyberta, zhang-etal-2021-need, trott2026large}. While previous rounds of BabyLM have focused on monolingual acquisition, the most recent iteration offers a multilingual track \citep{jumelet2025babybabellm}. 
In contrast to the goal of discovering data-efficient learning algorithms, here we adopt a ``controlled rearing'' approach, in which model architecture is held constant and training data, in the form of child-directed speech, are systematically manipulated \citep{frank2023openly}. In particular, we adopt a synthetic data approach introduced by \citet{eldan2023tinystories} and adapted to the developmental context by \citet{feng-etal-2024-child}. This approach has been used for systematic investigations of the impacts of training data on model performance. For example, \citet{kallini-etal-2024-mission} construct synthetic ``impossible'' languages by systematically altering English word order to probe what inputs were most challenging for SLMs (as representations of agnostic statistical learners). 

Several more recent investigations have also used the BabyLM approach to investigate multilingual learning, often in the context of second language learning \citep{oba-etal-2023-second,constantinescu-etal-2025-investigating,yadavalli-etal-2023-slabert,aoyama-schneider-2024-modeling, shen-etal-2024-bambino}. In one study, \citet{constantinescu-etal-2025-investigating} investigated whether sequential language exposure results in performance decrements (a ``critical period'' effect), a question also explored by \citet{plantinga2026lost,muller2025probing}. In contrast, we examine different structures of simultaneous exposure. Other work finds that parallel data improves translation more than code-switching \citep{shao-etal-2026-role}, although code-switched curricula can modestly improve grammaticality \citep{rooryck2026feeding}.
\citet{binyamin2026learningchilddirectedspeechtwolanguage} is the study closest to ours: they train SLMs on English-French corpora and study effects on syntactic and semantic tasks. While related, these studies differ from ours in several key respects: in particular, none test different exposure regimes (e.g., differentiating exposure by speaker; code-switching) and none use translated corpora to match exposure exactly across languages.

\section{Methods}
\label{headings}


\subsection{Training Data}\label{subsec:train_conds}
Our aim was to create a set of synthetic, matched corpora that instantiate monolingual and bilingual exposures, following \citet{feng-etal-2024-child}'s synthetic TinyDialogues corpus.\footnote{We provide a control using natural data in Section \ref{sec:addit}.} We began by using a commercial LLM (GPT-4/Gemini 3.1) to construct a diverse, synthetic dataset of 100M English (L1) words of dialogues. Each conversation involves two participants -- an adult speaker (mother or father) and a child (age 2, 5, 10, or 15). Prompt templates, generation details, example dialogues, and preprocessing statistics are provided in Appendices \ref{appendix:datacollection} and \ref{appendix:datapreprocessing}.

To construct parallel corpora, we used GPT-4 and Gemini 3.1 to translate the generated English (L1) dialogues into three second languages (L2s): Spanish, Chinese, and Dutch, which reflect varying linguistic relationships to English. English–Spanish bilingualism is extensively studied \citep{potter2025infants,kapantzoglou2021code,ibanez2010inhibitory,kaushanskaya2009bilingual,carlson2008bilingual}. Chinese provides a linguistically and orthographically distinct contrast, posing additional challenges for tokenization and evaluation \citep{song2026first,wang2024tokenization}. Dutch is more closely related to English and offers established evaluation resources \citep{brans-bloem-2026-multi}.

Code-switching is not random but instead reflects underlying linguistic constraints. Following prior work \citep{yoo-etal-2025-code-switching,wang-etal-2025-investigating-scaling}, we used two distinct methods to transform our parallel corpora into sentence- and word-level code-switching. For sentence-level code-switching, dialogues were segmented by punctuation and each sentence was probabilistically drawn from English or L2 corpora (avoiding runs of more than three sentences from one language). To construct word-level code-switching, we prompted GPT-4 to generate code-switched sentences given explicit structural constraints \citep{kuwanto2024linguistics,zeng2024leveraging,xie2025switchlingua,yoo-etal-2025-code-switching,wang-etal-2025-investigating-scaling}.

\subsection{Training Conditions}
We next used these four matched corpora for each L2 to create a series of controlled training conditions to test whether specific contrasts in training data alter performance. 
Specifically, we created the following conditions, as illustrated in Figure~\ref{fig:intro}:

\begin{itemize}
\item {\bf English Topline}. Models were trained on the full 100M corpus of English dialogues, serving as an upper bound on English performance. 
\item {\bf L2 Topline}. Models were trained on the full 100M L2-dialogue corpus.
\item {\bf English Baseline}. Models were trained on 50M words of English, either a {\bf random} selection, including both speakers, or a selection for a single speaker ({\bf speaker}). Results are averaged across speaker–language assignments to control for speaker-specific content biases.
\item {\bf Bilingual}. Models were trained on 50M words of English and 50M of L2, either a {\bf random} selection, in which either speaker may produce either language, or by {\bf speaker}, corresponding to ``one speaker, one language.''
\item {\bf Code-switching}. Models were trained on data in which English and L2 exposure alternated within either a single dialogue ({\bf sentence-level}) or a single sentence ({\bf word-level}). 
\end{itemize}


\subsection{Model Training}

We pretrained a 124M-parameter GPT-2 Small autoregressive language model \citep{radford2019language}, following prior ``controlled rearing'' work \citep{feng-etal-2024-child, kallini-etal-2024-mission, misra-mahowald-2024-language}. We additionally experimented with GPT-2 Mini (39M parameters) and GPT-2 Tiny (16M parameters). Models were trained for 20 epochs using Adam ($\beta=(0.9,0.999)$, $\epsilon=10^{-8}$), a learning rate of $10^{-4}$ with linear decay and no warmup, a maximum sequence length of 1024, and per-GPU batch sizes of 4--16. We ran three seeds (42, 0, 1) per condition, selecting checkpoints by lowest validation loss.

Tokenization poses challenges for multilingual training \citep{haslett-2025-tokenization,chang-etal-2024-multilinguality,wang2024tokenization}. Accordingly, we trained separate GPT-2 byte-level BPE tokenizers for each condition, increasing the maximum vocabulary from 52k to 80k to reduce excessive subword fragmentation in multilingual corpora \citep{rust-etal-2021-good, abagyan-etal-2026-one}. Actual vocabulary sizes varied by condition, with smaller monolingual corpora falling below 80k (see Appendix~\ref{appendix:token} for details).\footnote{ Vocabulary size also affects model parameter counts through the embedding matrix.}

\subsection{Evaluation}\label{sec:eval}
We evaluate three core dimensions of language learning: semantic knowledge, grammatical knowledge, and perplexity (PPL), following prior SLM evaluations \citep{constantinescu-etal-2025-investigating, feng-etal-2024-child, chen2026assessing}. Capturing fundamental linguistic competencies, these measures reflect the closely intertwined development of vocabulary and grammar in children, characterized as largely unifactorial \cite{brinchmann2019there}.


{\bf Perplexity (PPL)} is the exponentiated average negative log-likelihood over a sequence of tokens. It provides a broad measure of an LM’s predictive uncertainty, with lower PPL broadly reflecting stronger linguistic competence. We computed token-level PPL on English and L2 validation data, using a sliding window with a 512-token stride to handle sequences exceeding the context window.

{\bf Semantics.}
We evaluate semantic knowledge using a composite English word similarity benchmark \citep{zhuang2023visual} (see Appendix~\ref{appendix:evaluationdatapreprocess} for included datasets) and MultiSimLex for L2 and cross-lingual evaluation \citep{vulic-etal-2020-multi,brans-bloem-2026-multi}.  We extract word embeddings from each hidden layer, averaging across subword tokens, and report the best-layer Spearman correlation between cosine similarities and human judgments (higher is better).


{\bf Grammaticality.}
We evaluated models on Zorro \citep{huebner-etal-2021-babyberta}, an English benchmark evaluating grammatical competence, and MultiBliMP/ZhoBliMP \citep{jumelet-etal-2026-multiblimp,liu-etal-2026-systematic} for L2. Zorro tests 13 syntactic phenomena with minimal pairs of grammatical and ungrammatical sentences derived from child-directed speech.

Because vocabulary coverage and tokenization substantially affect performance, we filter evaluation items to ensure their words appear in the baseline training corpora and, for word similarity, are not fragmented by the tokenizer (Appendix~\ref{appendix:evaluationdatapreprocess}). 



\subsection{Additional Experiments}\label{sec:addit}

We conducted additional experiments in Spanish and Chinese to examine the effects of exposure proportions and resource constraints (RQ3). Beyond the standard 50--50 L1--L2 split, we tested ratios of 75--25, 25--75, 90--10, and 10--90 in the Bilingual (random) condition to reflect variation in bilingual exposure. To investigate whether bilingual effects emerge under more constrained conditions, we reduced model size from GPT-2 Small (124M) to Mini (39M) while holding training data fixed, and separately reduced training data to 20M words.

We also conducted a number of control experiments in Spanish: we further reduced model size to 16M parameters; we replaced Spanish data with randomly shuffled words and characters to test whether effects depend on linguistic structure; we replicated our conditions using naturalistic child-directed speech from CHILDES; and finally, to control for tokenization differences, we retrained models using a shared tokenizer trained on both English and L2 corpora.

\section{Results}

\subsection{Main Experiment}\label{sec:main_expt_results}

\begin{table}[t]
\centering
\small
\setlength{\tabcolsep}{2.5pt}
\begin{tabular}{@{}lcccc@{}}
\toprule
Model & English & Spanish & Chinese & Dutch \\
\midrule

English Topline
& $2.4{\pm}0.0$ & $>10^3$
& $>10^3$ & $>10^3$ \\

L2 Topline
& $>10^3$ & $2.6{\pm}0.0$
& $5.3{\pm}0.0$ & $2.8{\pm}0.0$ \\

\midrule

Baseline (speaker)
& $4.6{\pm}0.2$ & $>10^3$
& $>10^3$ & $>10^3$ \\

Baseline (random)
& $2.7{\pm}0.0$ & $>10^3$
& $>10^3$ & $>10^3$ \\

\midrule

Bilingual (speaker)
& $3.0{\pm}0.0$ & $3.4{\pm}0.0$
& $7.3{\pm}0.1$ & $3.5{\pm}0.0$ \\

Bilingual (random)
& $2.6{\pm}0.0$ & $2.8{\pm}0.0$
& $6.0{\pm}0.0$ & $3.0{\pm}0.0$ \\

\midrule

Code-switch (sent)
& $5.3{\pm}0.6$ & $6.1{\pm}0.6$
& $23.9{\pm}1.5$ & $4.6{\pm}0.1$ \\

Code-switch (word)
& $3.2{\pm}0.1$ & $3.7{\pm}0.0$
& $16.6{\pm}0.2$ & $3.9{\pm}0.0$ \\

\bottomrule
\end{tabular}
\caption{Perplexity (mean $\pm$ SD) across three seeds. Lower is better. English perplexity is reported for English-Spanish experiments (full results in Tables~\ref{tab:mainapp1}--\ref{tab:mainapp3}). Chinese perplexity is overall higher due to tokenization differences.  }
\label{tab:ppl}
\end{table}

\begin{figure*}[t]
\begin{center}
\includegraphics[width=\textwidth]{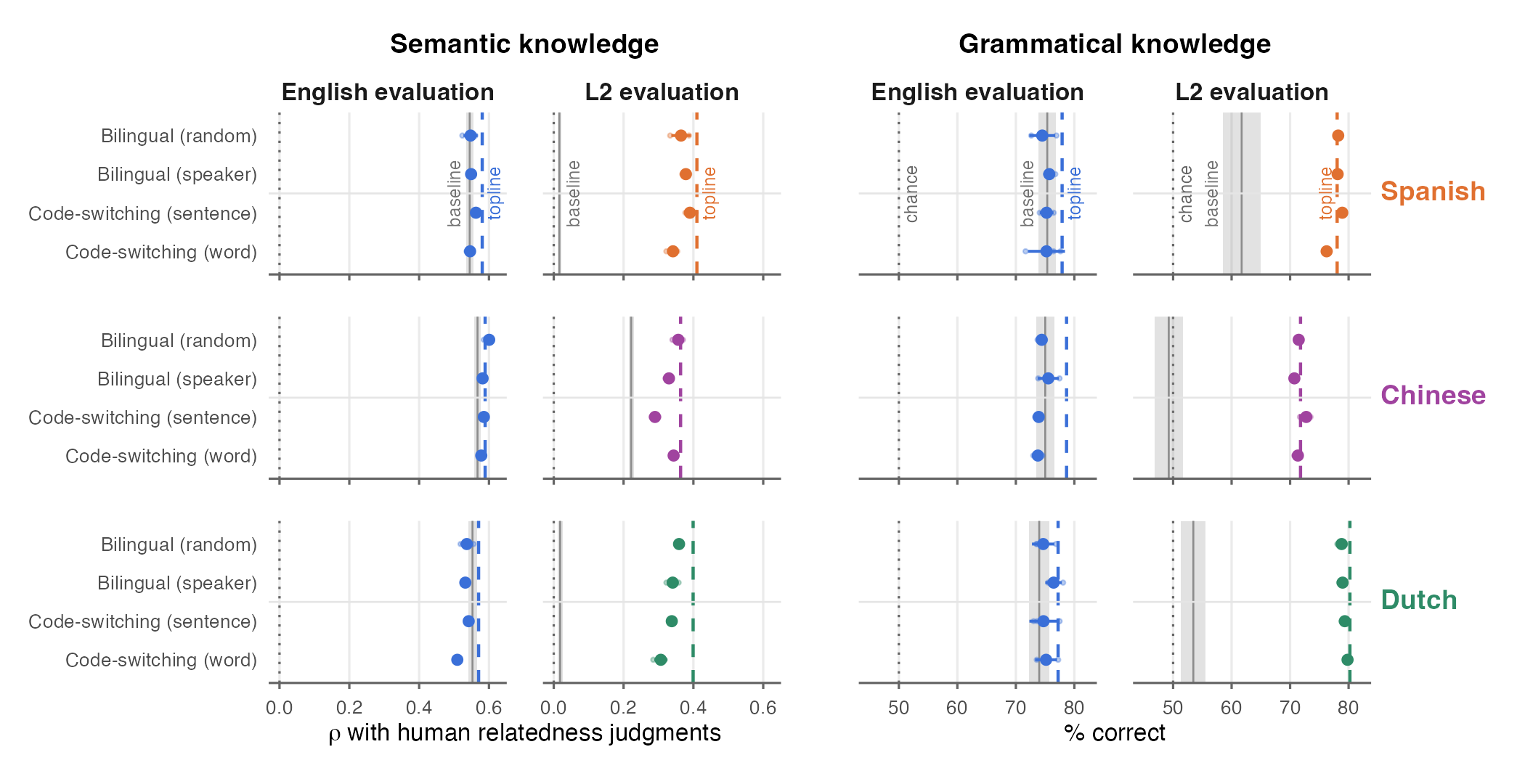}
\caption{Semantic and grammatical evaluation of GPT-2 (124M) across training conditions and languages. Dots show means across three seeds, with horizontal bars indicating ±1 SD. Shaded bands show English Baseline means ±1 SD. Baseline values are above zero for Chinese L2 semantic evaluation due to the presence of shared characters between words and for Spanish L2 grammatical evaluation due to length being confounded with correctness in Spanish MultiBLiMP. We report mean ± SD numerically in Tables~\ref{tab:mainapp1}--\ref{tab:mainapp3}, as well as cross-lingual word similarity.
\label{fig:main_results}}
\end{center}
\end{figure*}

Perplexity (PPL) values suggest that models effectively learned in all conditions where the training data matched the language of evaluation (Table~\ref{tab:ppl}). 
Multilingual models showed no ``confusion'': their English PPL was comparable to, and in some cases slightly lower than, English baselines, suggesting that additional L2 exposure provided a modest benefit. Code-switching models likewise learn both languages without evidence of English confusion despite mixed input. Sentence-level code-switching resulted in higher PPL than word-level switching and other multilingual conditions, likely because sentence-level code-switching aligns higher-level semantic topics but provides weaker local cues for predicting when a language shift will occur, whereas word-level code-switching provides explicit within-sentence constraints that make switching points more predictable.

Multilingual models showed little or no difference on English syntactic (Zorro) and semantic (WS) evaluations (Figure~\ref{fig:main_results}). 
Bilingual and code-switching models were within the English baseline's standard deviation and almost reached the topline, with limited differences between conditions. However, all bilingual and code-switching conditions show a clear benefit in L2 evaluations, where performance approached topline in both semantics and grammaticality. Appendix \ref{app:main} shows comparable results for cross-lingual WS. 

\subsection{Parametrically Varying Exposure}\label{sec:varying_exposure}

\begin{figure*}[t]
\begin{center}
\includegraphics[width=\textwidth]{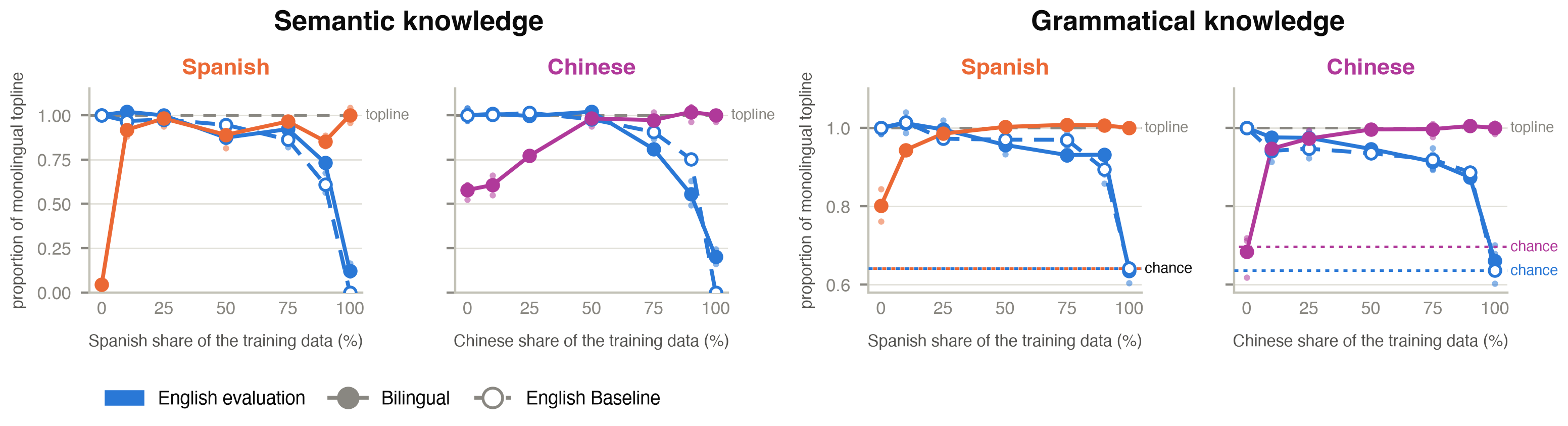}
\caption{Evaluation results for GPT-2 (124M) trained with varying language proportions in the Bilingual (random) condition. Scores are normalized by the corresponding monolingual topline for each evaluation language, so that 1.0 represents topline performance. Markers denote means over three seeds. 
\label{fig:varying_exposure_graph}
}
\end{center}
\end{figure*}

We next manipulated the percentage of English and L2 data within the same 100M-word corpus (for Spanish and Chinese only). Increasing the exposure ratio of English to L2 led to supralinear English performance, with little evidence of interference from multilingual data (Figure ~\ref{fig:varying_exposure_graph}). English Baseline performance largely overlapped with Bilingual performance, replicating the lack of difference between English baseline and Bilingual models in the first set of experiments. In L2, the pattern of learning is largely symmetrical: a small amount of exposure led to relatively high performance.

\subsection{Reduced Model and Data Scale}\label{sec:reduced}


We next test if these results hold for smaller data and model sizes. We provide English and Chinese performance in semantic and grammatical knowledge for models trained on 1/5 the data (20M words) and models of a smaller scale (GPT-2 Mini rather than GPT-2 Small; details in Appendix~\ref{app:reducedscale}). While decreasing model size did not decrease performance, reducing data exposure decreased performance across the board (Figures \ref{fig:gpt2mini} and \ref{fig:20M}). 
We further extended our Spanish experiments to an even smaller GPT-2 Tiny, observing only marginal performance decreases and similar relative trends across conditions (Figure~\ref{fig:manyscales}).
The gap between the English topline and other conditions widened with less data, but there was still no measurable difference in semantic or grammatical performance between models trained entirely in English (baseline) and those also exposed to an L2.


\subsection{The Role of Linguistic Structure}\label{sec:role}

Our experiments show no negative (or positive) effect of adding additional training data in an L2. Does this null effect depend on the L2 data being true natural language data, or are models able to handle even unstructured L2 data? To test this question, we first shuffled the Spanish corpus (randomizing word order) while holding English exposure fixed (Shuffled Spanish). This manipulation resulted in no difference in English performance (Table \ref{tab:es-controls}). We also  replaced the Spanish portion with randomly generated characters (Random Characters). When the tokenizer was retrained for each condition, performance decreased substantially in the Random Characters condition because the random characters expanded the vocabulary and increased token fragmentation. However, when we controlled for tokenizer effects by using the Bilingual (random) tokenizer across conditions, adding even random data did not result in major decrements in English performance. 


\subsection{Extension to Naturalistic Data}\label{sec:natural}

\begin{table*}[t]
\centering
\small
\setlength{\tabcolsep}{3.5pt}

\resizebox{\textwidth}{!}{%
\begin{tabular}{lccc c c cc}
\toprule
& \multicolumn{3}{c}{\textbf{Semantic Knowledge}}
& \multicolumn{2}{c}{\textbf{Grammatical Knowledge}}
& \multicolumn{2}{c}{\textbf{Perplexity}} \\
\cmidrule(lr){2-4}
\cmidrule(lr){5-6}
\cmidrule(lr){7-8}

\textbf{Model}
& \textbf{EN}
& \textbf{ES}
& \textbf{EN-ES}
& \textbf{EN}
& \textbf{ES}
& \textbf{EN}
& \textbf{ES} \\
\midrule

CHILDES (English Topline)
& $0.39 \pm 0.01$
& $0.05 \pm 0.01$
& $0.11 \pm 0.00$
& $74.93 \pm 1.59$
& $62.17 \pm 1.05$
& $3.79 \pm 0.00$
& $>10^{3}$ \\

CHILDES (Spanish Topline)
& $0.01 \pm 0.01$
& $0.31 \pm 0.02$
& $0.05 \pm 0.01$
& $53.07 \pm 3.10$
& $75.65 \pm 0.46$
& $26.91 \pm 0.35$
& $3.74 \pm 0.01$ \\

\midrule 

CHILDES (English Baseline)
& $0.30 \pm 0.00$
& $0.05 \pm 0.02$
& $0.10 \pm 0.01$
& $72.09 \pm 1.38$
& $61.41 \pm 1.27$
& $4.10 \pm 0.01$
& $>10^{3}$ \\

\midrule

CHILDES (Bilingual)
& $0.31 \pm 0.02$
& $0.25 \pm 0.03$
& $0.18 \pm 0.02$
& $73.28 \pm 1.50$
& $76.23 \pm 1.06$
& $3.99 \pm 0.01$
& $3.95 \pm 0.01$ \\

CHILDES (Code-switching)
& $0.36 \pm 0.01$
& $0.26 \pm 0.02$
& $0.29 \pm 0.01$
& $73.62 \pm 1.44$
& $75.90 \pm 0.18$
& $4.81 \pm 0.13$
& $4.88 \pm 0.16$ \\

\bottomrule
\end{tabular}%
}
\caption{
Semantic, grammatical, and perplexity results for models trained on CHILDES-derived corpora (29M words). EN-ES denotes cross-lingual word similarity. Values indicate mean $\pm$ standard deviation across three seeds.}
\label{tab:spanish-results}
\end{table*}

Do the synthetic data used in our primary experiments provide a reliable proxy for human language acquisition? To address this concern, we replicate our experiments using English data from CHILDES \cite{macwhinney2014childes}, 29M words of naturally occurring, transcribed interactions between children and caregivers. Following our corpus-matching strategy, we translated the CHILDES corpus into Spanish using the same procedure described in Section \ref{appendix:datacollection}, aligned the parallel English and Spanish sentences, and constructed a matched code-switching (sentence) condition. This procedure necessarily introduces some artificiality because the Spanish data are machine-translated rather than naturally occurring Spanish child-directed speech. 
We then train models using the same training procedure and apply the same evaluation and preprocessing pipeline as in our main experiments. 
Results were quite similar: the English baseline and multilingual models perform similarly within error on English evaluations, while exposure to Spanish substantially improves Spanish performance (Table~\ref{tab:spanish-results}). 




\subsection{Matching Tokenization Effects}\label{sec:match}

In our experiments, model performance should be understood as a property of the tokenizer-model system rather than of the model in isolation. To test whether our conclusions are sensitive to tokenizer choice, we repeat the main experiments using a single shared tokenizer trained jointly on the full 100M-word English and 100M-word Spanish parallel corpora, with the same 80k vocabulary size. This tokenizer is therefore exposed to the complete lexical distribution that may appear in any experimental condition. We retrained each model from scratch while holding this tokenizer fixed across conditions, noting that a single joint tokenizer is ``implicitly bilingual'' due to its use of a multilingual tokenization scheme. However, using this joint tokenizer, both the relative trends across conditions and the absolute scores were nearly unchanged from the main experiments (Table~\ref{tab:joint_tokenizer}). 

\subsection{Qualitative Analysis}\label{sec:modelinterp}

\begin{figure*}[t]
\centering

\begin{subfigure}[t]{0.3\textwidth}
    \centering
    \includegraphics[width=\linewidth]{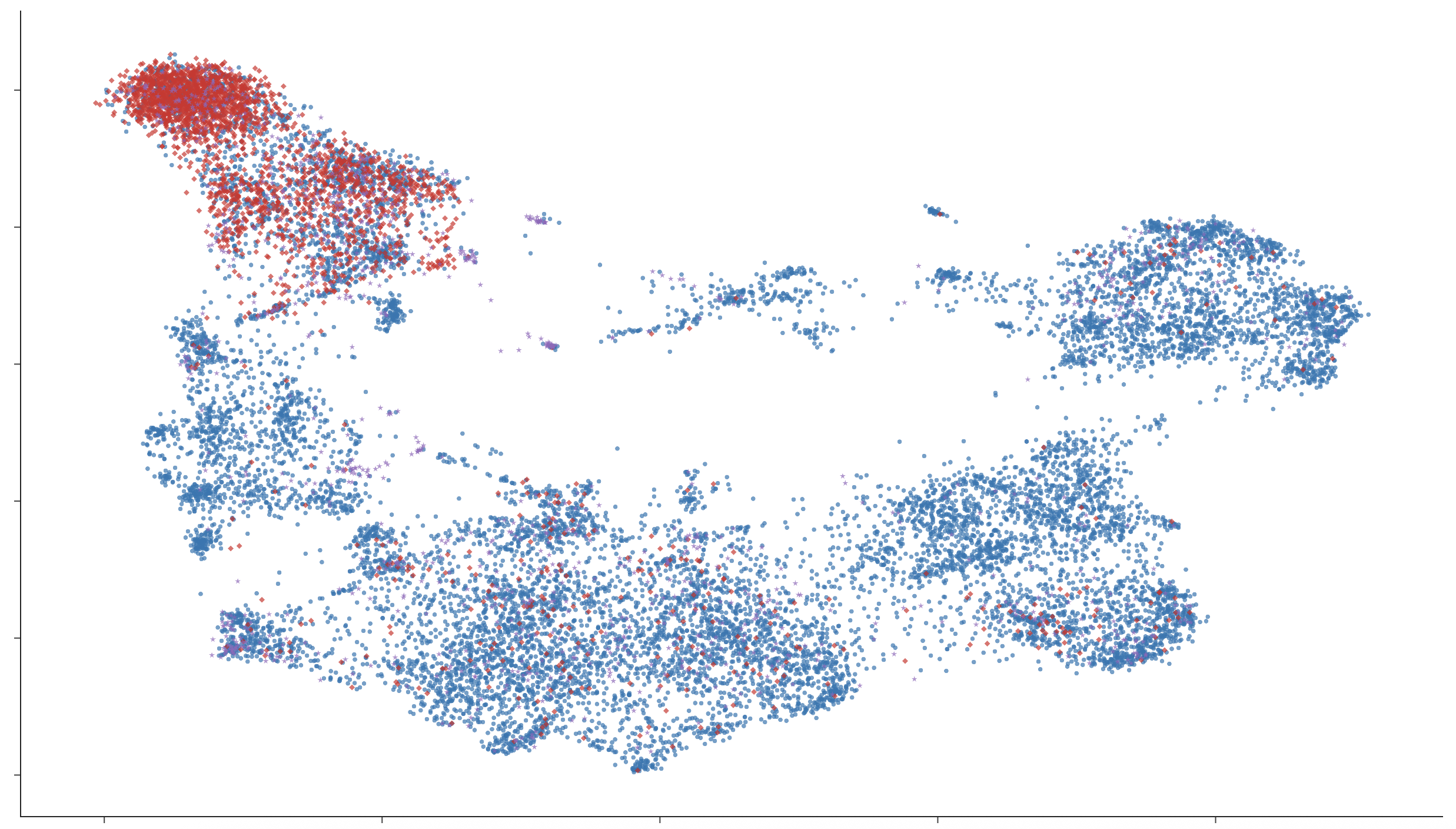}
    \caption{English Baseline (random)}
\end{subfigure}
\hfill
\begin{subfigure}[t]{0.3\textwidth}
    \centering
    \includegraphics[width=\linewidth]{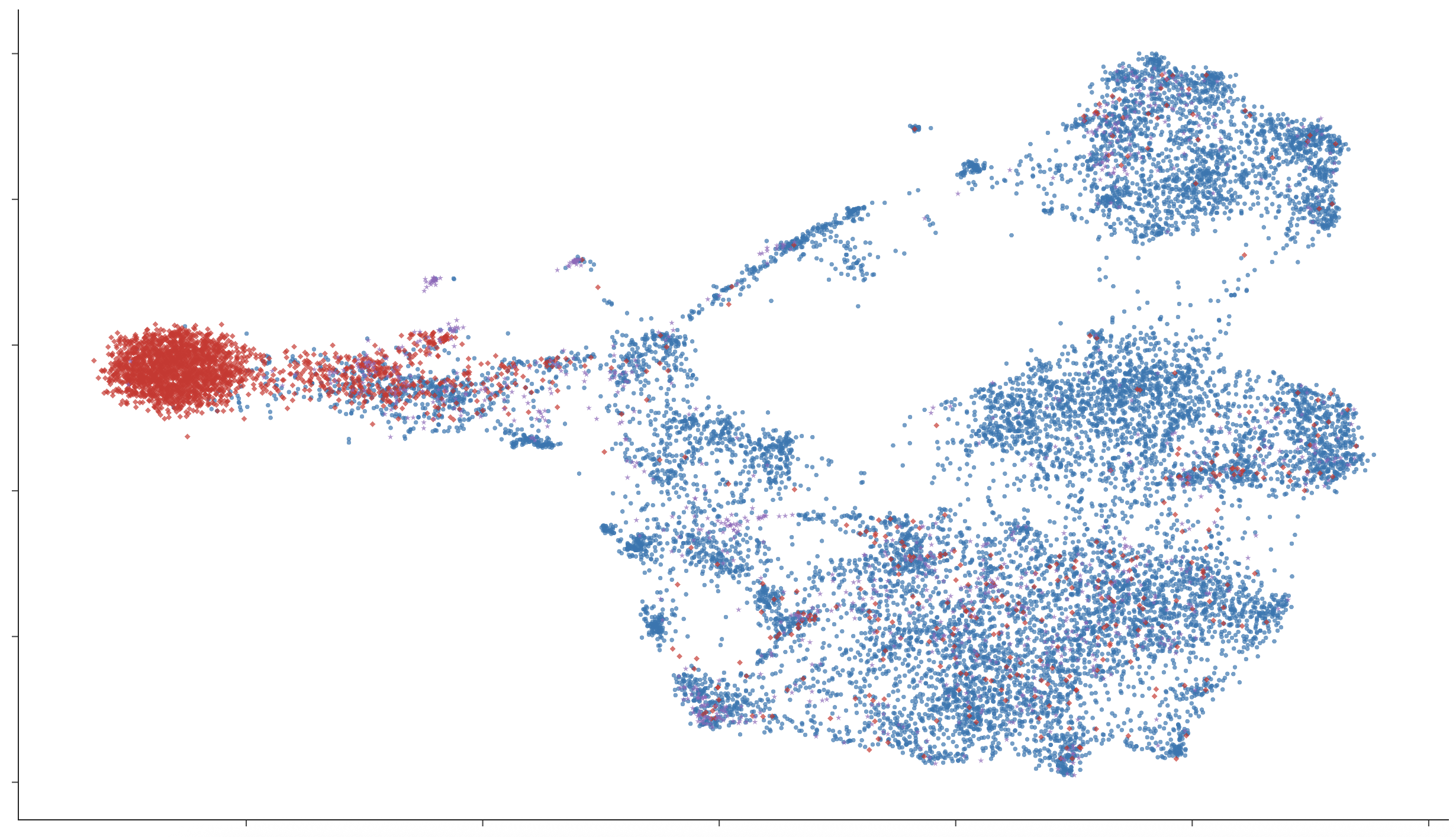}
    \caption{English Topline}
\end{subfigure}
\hfill
\begin{subfigure}[t]{0.3\textwidth}
    \centering
    \includegraphics[width=\linewidth]{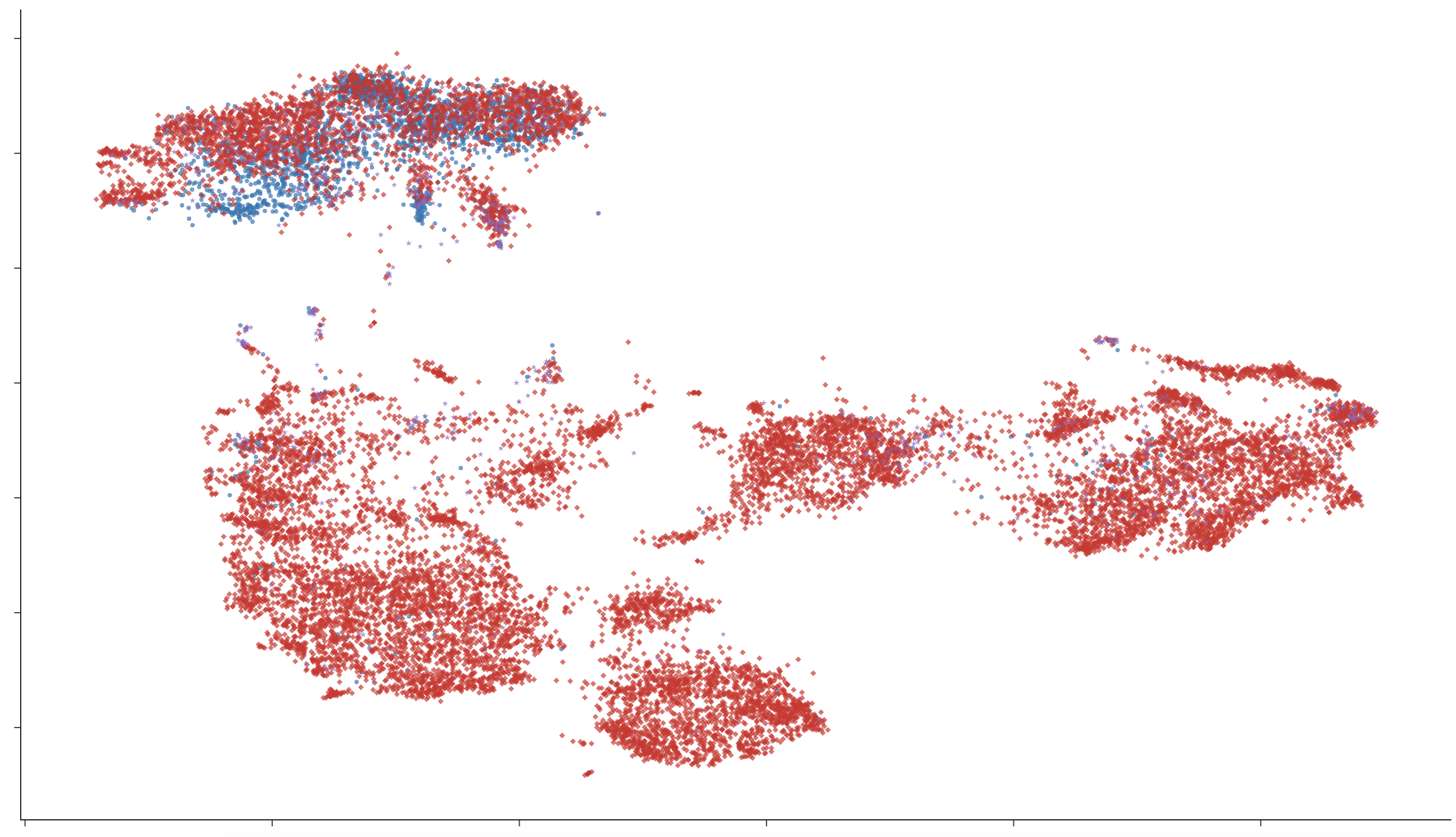}
    \caption{Spanish Topline}
\end{subfigure}

\vspace{0.5em}

\begin{subfigure}[t]{0.3\textwidth}
    \centering
    \includegraphics[width=\linewidth]{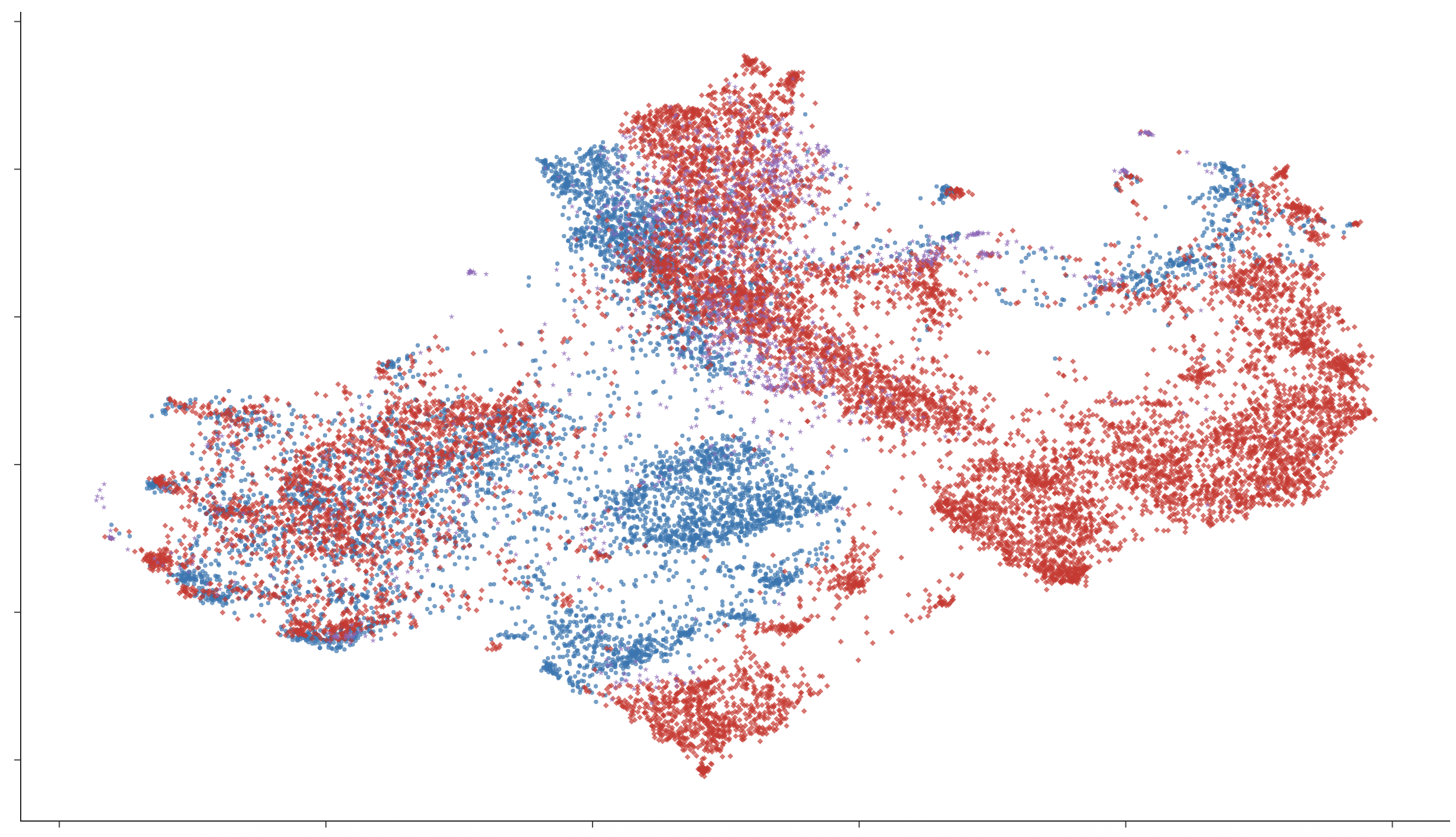}
    \caption{Bilingual (random)}
\end{subfigure}
\hfill
\begin{subfigure}[t]{0.3\textwidth}
    \centering
    \includegraphics[width=\linewidth]{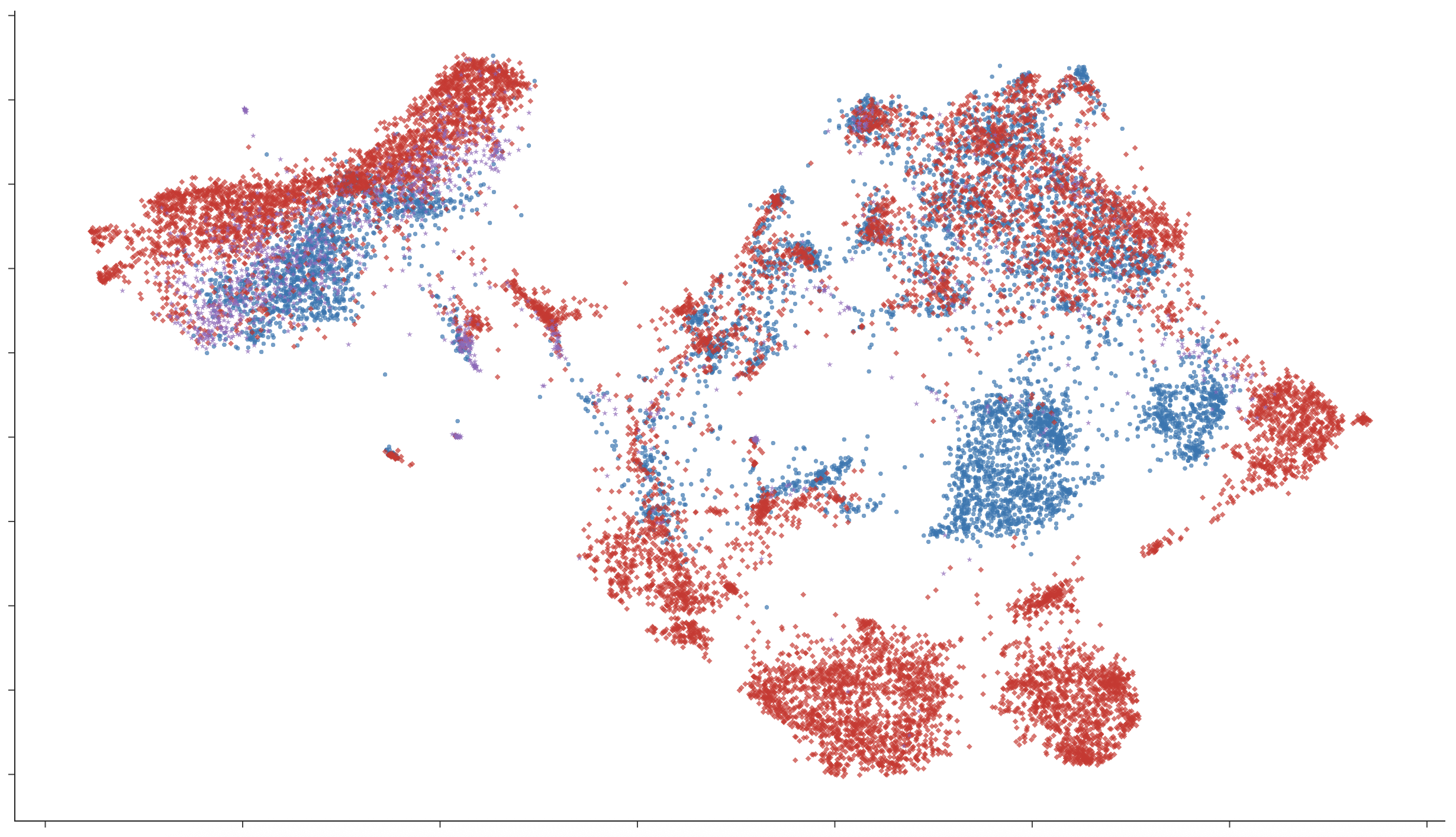}
    \caption{Code-switching (sentence)}
\end{subfigure}
\hfill
\begin{subfigure}[t]{0.3\textwidth}
    \centering
    \includegraphics[width=\linewidth]{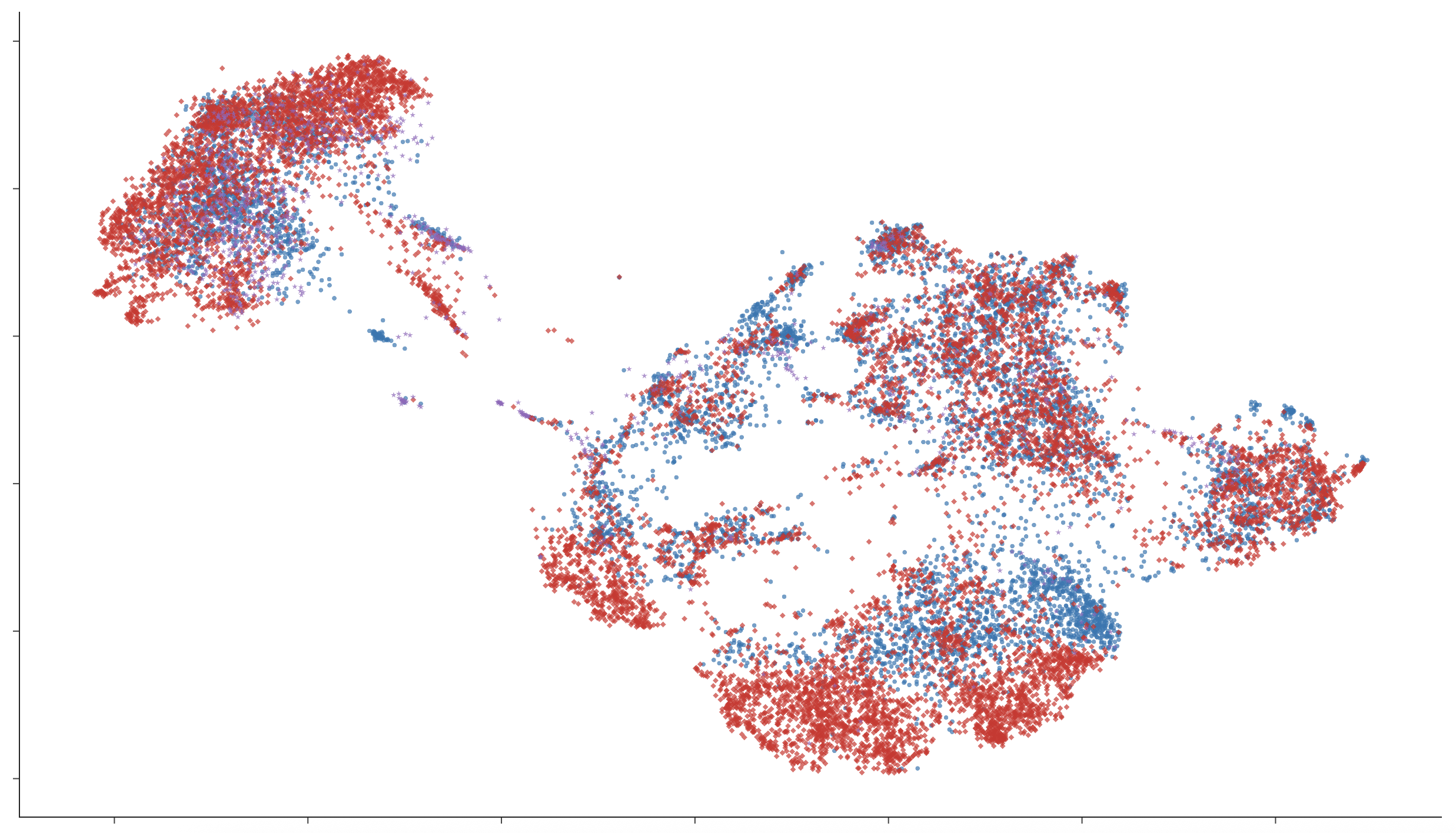}
    \caption{Code-switching (word)}
\end{subfigure}

\caption{Two-dimensional visualizations of the trained GPT-2 (124M) token embedding space, with  English, Spanish, and shared tokens in blue, red, and purple respectively. 
}
\label{fig:embedding_grid}
\end{figure*}


To examine the structure of learned lexical representations, we visualized token embeddings from English-Spanish GPT-2 models trained under different conditions (Figure~\ref{fig:embedding_grid}). We extracted embeddings for a subset of 20k tokens and projected them into two dimensions using Uniform Manifold Approximation and Projection to preserve local neighborhood structure \citep{McInnes2018}. Tokens were labeled by language based on their relative frequencies in the training corpora (details in Appendix~\ref{appendix:viz}).
Despite the lack of confusion or interference effects, the lexical space of multilingual models appears inherently mixed rather than fully separated, and increasing the granularity of bilingual exposure progressively integrates the representational space. While monolingual baselines and toplines (\ref{fig:embedding_grid}a--c) show monolingual clusters, the Bilingual conditions (\ref{fig:embedding_grid}d–f) display an even distribution of both languages in a hybrid structure, with increasing overlap in code-switching. 

\section{Discussion}
Multilingualism raises parallel questions in human language acquisition and language modeling about how learners acquire multiple input distributions across exposure structures, model capacities, and data scales. Using language models as controlled statistical learners, we systematically manipulated these exposure structures to investigate their effects on language acquisition.
We found that multilingual training neither confers advantages nor induces confusion relative to monolingual training (\textbf{RQ1}); instead, bilingual models simply gained knowledge of a second language. In addition, performance was relatively similar across quite different multilingual exposure regimes (\textbf{RQ2}). 

These findings persisted across languages, model sizes, and training data scales (\textbf{RQ3}), as well as in experiments using naturalistic child-caregiver interactions. Across conditions, performance was primarily determined by the amount of training data available in each language, following an approximately logarithmic relationship consistent with prior scaling-law research \citep{hoffmann2022training}. Even replacing L2 input with shuffled words or random characters produced little change in L1 performance, suggesting that additional input need not even constitute a coherent second language to avoid interference. The most consequential differences instead arose from tokenization, which altered how linguistic input was represented and how much effective training signal models received.



Although multilingual learning raises parallel questions in humans and LMs, their learning mechanisms differ substantially. Perhaps most obviously, SLMs process raw text and lack access to the phonological and prosodic differences that make two languages easily distinguishable. Moreover, the Bilingual (speaker) conditions in our experiments used simple text labels to distinguish speakers, while human learners have access to all the myriad of differences between  caregivers.  
Our contributions come from an analysis of the statistical properties of multilingual input \emph{in the absence} of these naturalistic features \cite{trott2026large}. We conclude that differences in performance between mono- and multilingual learners do not arise due to statistical properties of language input -- though they may still arise in human learners for other reasons! This finding was not a foregone conclusion: indeed, several studies on bilingualism in LMs find significant confusion and L1 attrition \citep{constantinescu-etal-2025-investigating, oba-etal-2023-second}.

\section{Conclusion}

We find little evidence that bilingual training induces confusion or degrades first-language performance in SLMs. While the mechanisms underlying human and model learning may differ, our results nonetheless provide evidence against a strong version of the confusion hypothesis: multilingual data do not \emph{necessarily} lead to any kind of interference. More broadly, by systematically manipulating the structure of training data while holding the learner constant, our experiments provide a causal framework for studying how multilingual environments shape learning outcomes. 

\section{Limitations}
While small language models (SLMs) provide a controlled framework for studying language exposure, they differ substantially from human learners, limiting the extent to which our findings can be generalized to human language acquisition. In particular, our models learn from text rather than speech and lack the social, perceptual, and interactive context of human learning. Although we show in Section~\ref{sec:natural} that our findings extend to naturalistic child-directed speech, our primarily synthetic training data remain an imperfect approximation of real-world multilingual exposure, which varies considerably between individuals and social contexts, particularly in patterns of code-switching \citep{de2019not}.

Tokenization remains an imperfect representation of linguistic input \citep{wang2024tokenization,haslett-2025-tokenization,nair-resnik-2023-words}, and benchmark performance can be sensitive to token fragmentation, sequence length, and other evaluation artifacts \citep{rust-etal-2021-good,liu-etal-2026-systematic}. Moreover, many downstream tasks yield near-chance performance at this model scale \cite{choshen2026babylm, feng2026baby}, while fine-tuning introduces additional training that may obscure the effects of controlled pretraining exposure \cite{mosbach-etal-2020-interplay-fine}. Our evaluations are therefore limited to a subset of semantic and grammatical competencies that can be assessed without fine-tuning. The availability of such benchmarks, particularly for lower-resource languages, further restricts the languages we can study (e.g., Chinese and Dutch remain the primary non-English languages featured in BabyLM benchmarks \citep{choshen2026babylm}). These limitations leave open whether our findings generalize to broader model capabilities and other languages.
\section{Acknowledgements}
We gratefully acknowledge NVIDIA's Academic Grant Program and Modal for providing a portion of the compute resources that enabled this work. We also thank Alvin W.M. Tan, Virgina Marchman, and other members of Stanford's Language \& Cognition (LangCog) Lab for valuable comments and feedback.\\

\bibliography{anthology}

@book{macwhinney2014childes,
  title={The CHILDES project: Tools for analyzing talk, Volume I: Transcription format and programs},
  author={MacWhinney, Brian},
  year={2014},
  publisher={Psychology Press}
}

@inproceedings{hu2025production,
  author    = {Hu, Jennifer and Tan, Alvin Wei Ming and Feng, Steven Y. and Frank, Michael C.},
  title     = {Language production is harder than comprehension for children and language models},
  booktitle = {Proceedings of the Annual Meeting of the Cognitive Science Society},
  volume    = {47},
  year      = {2025},
  url       = {https://escholarship.org/uc/item/5rz8b9jg}
}

@article{song2026first,
  title={The First ChineseBabyLM Challenge: training data-efficient and cognitively plausible language models for Chinese},
  author={Song, Siyuan and Qian, Zhiheng and Zhang, Yunhao and He, Linyang and Ji, Xiaozhe and Lin, Yingxin and Zhu, Hongao and Shao, Chongtian and Lang, Chuhan and Li, Luan and others},
  journal={arXiv preprint arXiv:2607.10745},
  year={2026}
}

@inproceedings{brans-bloem-2026-multi,
    title = "Multi-{S}im{L}ex for {D}utch: Benchmarking Embedding- and Prompt-Based Model Performance on Semantic Similarity",
    author = "Brans, Lizzy  and
      Bloem, Jelke",
    editor = "Piperidis, Stelios  and
      Bel, N{\'u}ria  and
      van den Heuvel, Henk  and
      Ide, Nancy  and
      Krek, Simon  and
      Toral, Antonio",
    booktitle = "Proceedings of the Fifteenth Language Resources and Evaluation Conference",
    month = may,
    year = "2026",
    address = "Palma de Mallorca, Spain",
    publisher = "ELRA Language Resource Association",
    url = "https://aclanthology.org/2026.lrec-1.380/",
    doi = "10.63317/2q9dcx9cvnu9",
    pages = "4846--4860"
}

@article{choshen2026babylm,
  title={BabyLM Turns 4 and Goes Multilingual: Call for Papers for the 2026 BabyLM Workshop},
  author={Choshen, Leshem and Cotterell, Ryan and Gul, Mustafa Omer and Jumelet, Jaap and Linzen, Tal and Mueller, Aaron and Salhan, Suchir and Shah, Raj Sanjay and Warstadt, Alex and Wilcox, Ethan Gotlieb},
  journal={arXiv preprint arXiv:2602.20092},
  year={2026}
}

@article{haslett-2025-tokenization,
    title = "Tokenization Changes Meaning in Large Language Models: Evidence from {C}hinese",
    author = "Haslett, David A.",
    journal = "Computational Linguistics",
    volume = "51",
    number = "3",
    month = sep,
    year = "2025",
    address = "Cambridge, MA",
    publisher = "MIT Press",
    url = "https://aclanthology.org/2025.cl-3.3/",
    doi = "10.1162/coli_a_00557",
    pages = "785--814"
}

@article{wang2024tokenization,
  title={Tokenization matters! degrading large language models through challenging their tokenization},
  author={Wang, Dixuan and Li, Yanda and Jiang, Junyuan and Ding, Zepeng and Luo, Ziqin and Jiang, Guochao and Liang, Jiaqing and Yang, Deqing},
  journal={arXiv preprint arXiv:2405.17067},
  year={2024}
}

@article{trott2026large,
  title={Large Language Models as Distributional Baselines for Language Tasks},
  author={Trott, Sean and Michaelov, James and Jones, Cameron and Chang, Tyler and Bergen, Benjamin},
  journal={Open Mind: Discoveries in Cognitive Science},
  volume={10},
  pages={1213},
  year={2026}
}

@inproceedings{muller2025probing,
  title={Probing for experience-driven critical period effects in a large language model},
  author={Muller, Dyana and Li Lin, Koquiun and Kemp, Charles and Mollica, Francis},
  booktitle={Proceedings of the Annual Meeting of the Cognitive Science Society},
  volume={47},
  year={2025}
}

@article{arredondo2026missing,
  title={Missing the mix: Perceiving code-switches as a mechanism for bilingual adaptation},
  author={Arredondo, Maria M and Castro, Stephanie},
  journal={Child development perspectives},
  volume={20},
  number={2},
  pages={105--114},
  year={2026},
  publisher={Oxford University Press US}
}

@article{bialystok2017bilingual,
  title={The bilingual adaptation: How minds accommodate experience.},
  author={Bialystok, Ellen},
  journal={Psychological bulletin},
  volume={143},
  number={3},
  pages={233},
  year={2017},
  publisher={American Psychological Association}
}

@article{de2019not,
  title={Not all bilinguals are the same: A call for more detailed assessments and descriptions of bilingual experiences},
  author={De Bruin, Angela},
  journal={Behavioral Sciences},
  volume={9},
  number={3},
  pages={33},
  year={2019},
  publisher={MDPI}
}

@article{vulic-etal-2020-multi,
    title = "Multi-{S}im{L}ex: A Large-Scale Evaluation of Multilingual and Crosslingual Lexical Semantic Similarity",
    author = "Vuli{\'c}, Ivan  and
      Baker, Simon  and
      Ponti, Edoardo Maria  and
      Petti, Ulla  and
      Leviant, Ira  and
      Wing, Kelly  and
      Majewska, Olga  and
      Bar, Eden  and
      Malone, Matt  and
      Poibeau, Thierry  and
      Reichart, Roi  and
      Korhonen, Anna",
    journal = "Computational Linguistics",
    volume = "46",
    number = "4",
    month = dec,
    year = "2020",
    url = "https://aclanthology.org/2020.cl-4.5/",
    doi = "10.1162/coli_a_00391",
    pages = "847--897"
}

@article{jumelet-etal-2026-multiblimp,
    title = "{M}ulti{BL}i{MP} 1.0: A Massively Multilingual Benchmark of Linguistic Minimal Pairs",
    author = "Jumelet, Jaap  and
      Weissweiler, Leonie  and
      Nivre, Joakim  and
      Bisazza, Arianna",
    journal = "Transactions of the Association for Computational Linguistics",
    volume = "14",
    year = "2026",
    address = "Cambridge, MA",
    publisher = "MIT Press",
    url = "https://aclanthology.org/2026.tacl-1.10/",
    doi = "10.1162/tacl.a.600",
    pages = "193--216"
}

@article{liu-etal-2026-systematic,
    title = "A Systematic Assessment of Language Models with Linguistic Minimal Pairs in {C}hinese",
    author = "Liu, Yikang  and
      Shen, Yeting  and
      Zhu, Hongao  and
      Xu, Lilong  and
      Qian, Zhiheng  and
      Song, Siyuan  and
      Zhang, Kejia  and
      Tang, Jialong  and
      Zhang, Pei  and
      Yang, Baosong  and
      Wang, Rui  and
      Hu, Hai",
    journal = "Transactions of the Association for Computational Linguistics",
    volume = "14",
    year = "2026",
    address = "Cambridge, MA",
    publisher = "MIT Press",
    url = "https://aclanthology.org/2026.tacl-1.34/",
    doi = "10.1162/tacl.a.648",
    pages = "755--771"
}

@article{yang2026apples,
  title={Apples to Apples? Towards Comparable Crosslingual Language Model Evaluation},
  author={Yang, Xiulin and Wilcox, Ethan Gotlieb and Arnett, Catherine},
  journal={arXiv preprint arXiv:2608.25089},
  year={2026}
}

@article{plantinga2026lost,
  title={Lost but not erased: Finding traces of a forgotten language in neural speech models},
  author={Plantinga, Peter and Moore, Charlotte and Donhauser, Peter W and Byers-Heinlein, Krista and Klein, Denise},
  journal={arXiv preprint arXiv:2608.25976},
  year={2026}
}

@article{rooryck2026feeding,
  title={Feeding BabyLMs Macaroni: Code-Switching Curricula Cause Cross-Lingual Convergence},
  author={Rooryck, Dries and Cai, Alex and Belinkov, Yonatan and Alvarez-Melis, David and Brantley, Kiant{\'e}},
  journal={arXiv preprint arXiv:2609.30535},
  year={2026}
}

@inproceedings{shao-etal-2026-role,
    title = "The Role of Mixed-Language Documents for Multilingual Large Language Model Pretraining",
    author = "Shao, Jiandong  and
      Tang, Raphael  and
      Zhang, Crystina  and
      Sevegnani, Karin  and
      Stenetorp, Pontus  and
      Yang, Jianfei  and
      Lu, Yao",
    editor = "Liakata, Maria  and
      Moreira, Viviane P.  and
      Zhang, Jiajun  and
      Jurgens, David",
    booktitle = "Proceedings of the 64th Annual Meeting of the {A}ssociation for {C}omputational {L}inguistics (Volume 1: Long Papers)",
    month = jul,
    year = "2026",
    address = "San Diego, California, United States",
    publisher = "Association for Computational Linguistics",
    url = "https://aclanthology.org/2026.acl-long.1706/",
    doi = "10.18653/v1/2026.acl-long.1706",
    pages = "36807--36818",
    ISBN = "979-8-89176-390-6"
}

@inproceedings{nair-resnik-2023-words,
    title = "Words, Subwords, and Morphemes: What Really Matters in the Surprisal-Reading Time Relationship?",
    author = "Nair, Sathvik  and
      Resnik, Philip",
    editor = "Bouamor, Houda  and
      Pino, Juan  and
      Bali, Kalika",
    booktitle = "Findings of the Association for Computational Linguistics: EMNLP 2023",
    month = dec,
    year = "2023",
    address = "Singapore",
    publisher = "Association for Computational Linguistics",
    url = "https://aclanthology.org/2023.findings-emnlp.752/",
    doi = "10.18653/v1/2023.findings-emnlp.752",
    pages = "11251--11260"
}

@inproceedings{abagyan-etal-2026-one,
    title = "One Tokenizer To Rule Them All: Emergent Language Plasticity via Multilingual Tokenizers",
    author = {Abagyan, Diana  and
      Salamanca, Alejandro R.  and
      Cruz-Salinas, Andres Felipe  and
      Cao, Kris  and
      Lin, Hangyu  and
      Locatelli, Acyr  and
      Fadaee, Marzieh  and
      {\"U}st{\"u}n, Ahmet  and
      Hooker, Sara},
    editor = "Liakata, Maria  and
      Moreira, Viviane P.  and
      Zhang, Jiajun  and
      Jurgens, David",
    booktitle = "Proceedings of the 64th Annual Meeting of the {A}ssociation for {C}omputational {L}inguistics (Volume 1: Long Papers)",
    month = jul,
    year = "2026",
    address = "San Diego, California, United States",
    publisher = "Association for Computational Linguistics",
    url = "https://aclanthology.org/2026.acl-long.141/",
    doi = "10.18653/v1/2026.acl-long.141",
    pages = "3118--3136",
    ISBN = "979-8-89176-390-6"
}

@inproceedings{papadimitriou-etal-2023-multilingual,
    title = "Multilingual {BERT} has an accent: Evaluating {E}nglish influences on fluency in multilingual models",
    author = "Papadimitriou, Isabel  and
      Lopez, Kezia  and
      Jurafsky, Dan",
    editor = "Vlachos, Andreas  and
      Augenstein, Isabelle",
    booktitle = "Findings of the Association for Computational Linguistics: EACL 2023",
    month = may,
    year = "2023",
    address = "Dubrovnik, Croatia",
    publisher = "Association for Computational Linguistics",
    url = "https://aclanthology.org/2023.findings-eacl.89/",
    doi = "10.18653/v1/2023.findings-eacl.89",
    pages = "1194--1200"
}

@inproceedings{shen-etal-2024-bambino,
    title = "{BAMBINO}-{LM}: (Bilingual-)Human-Inspired Continual Pre-training of {B}aby{LM}",
    author = "Shen, Zhewen  and
      Joshi, Aditya  and
      Chen, Ruey-Cheng",
    editor = "Kuribayashi, Tatsuki  and
      Rambelli, Giulia  and
      Takmaz, Ece  and
      Wicke, Philipp  and
      Oseki, Yohei",
    booktitle = "Proceedings of the Workshop on Cognitive Modeling and Computational Linguistics",
    month = aug,
    year = "2024",
    address = "Bangkok, Thailand",
    publisher = "Association for Computational Linguistics",
    url = "https://aclanthology.org/2024.cmcl-1.1/",
    doi = "10.18653/v1/2024.cmcl-1.1",
    pages = "1--7"
}

@inproceedings{aoyama-schneider-2024-modeling,
    title = "Modeling Nonnative Sentence Processing with {L}2 Language Models",
    author = "Aoyama, Tatsuya  and
      Schneider, Nathan",
    editor = "Al-Onaizan, Yaser  and
      Bansal, Mohit  and
      Chen, Yun-Nung",
    booktitle = "Proceedings of the 2024 Conference on Empirical Methods in Natural Language Processing",
    month = nov,
    year = "2024",
    address = "Miami, Florida, USA",
    publisher = "Association for Computational Linguistics",
    url = "https://aclanthology.org/2024.emnlp-main.283/",
    doi = "10.18653/v1/2024.emnlp-main.283",
    pages = "4927--4940"
}

@inproceedings{yadavalli-etal-2023-slabert,
    title = "{SLABERT} Talk Pretty One Day: Modeling Second Language Acquisition with {BERT}",
    author = "Yadavalli, Aditya  and
      Yadavalli, Alekhya  and
      Tobin, Vera",
    editor = "Rogers, Anna  and
      Boyd-Graber, Jordan  and
      Okazaki, Naoaki",
    booktitle = "Proceedings of the 61st Annual Meeting of the Association for Computational Linguistics (Volume 1: Long Papers)",
    month = jul,
    year = "2023",
    address = "Toronto, Canada",
    publisher = "Association for Computational Linguistics",
    url = "https://aclanthology.org/2023.acl-long.657/",
    doi = "10.18653/v1/2023.acl-long.657",
    pages = "11763--11777"
}

@misc{binyamin2026learningchilddirectedspeechtwolanguage,
      title={Learning from Child-Directed Speech in Two-Language Scenarios: A French-English Case Study}, 
      author={Liel Binyamin and Elior Sulem},
      year={2026},
      eprint={2603.12906},
      archivePrefix={arXiv},
      primaryClass={cs.CL},
      url={https://arxiv.org/abs/2603.12906}, 
}

@article{hoffmann2022training,
  title={Training compute-optimal large language models},
  author={Hoffmann, Jordan and Borgeaud, Sebastian and Mensch, Arthur and Buchatskaya, Elena and Cai, Trevor and Rutherford, Eliza and Casas, DDL and Hendricks, Lisa Anne and Welbl, Johannes and Clark, Aidan and others},
  journal={arXiv preprint arXiv:2203.15556},
  volume={10},
  year={2022}
}

@article{byers2017bilingual,
  title={Bilingual infants control their languages as they listen},
  author={Byers-Heinlein, Krista and Morin-Lessard, Elizabeth and Lew-Williams, Casey},
  journal={Proceedings of the National Academy of Sciences},
  volume={114},
  number={34},
  pages={9032--9037},
  year={2017},
  publisher={National Academy of Sciences}
}

@article{guiberson2013bilingual,
  title={Bilingual myth-busters series language confusion in bilingual children},
  author={Guiberson, Mark},
  journal={Perspectives on Communication Disorders and Sciences in Culturally and Linguistically Diverse (CLD) Populations},
  volume={20},
  number={1},
  pages={5--14},
  year={2013},
  publisher={American Speech-Language-Hearing Association}
}

@article{dick2019no,
  title={No evidence for a bilingual executive function advantage in the ABCD study},
  author={Dick, Anthony Steven and Garcia, Nelcida L and Pruden, Shannon M and Thompson, Wesley K and Hawes, Samuel W and Sutherland, Matthew T and Riedel, Michael C and Laird, Angela R and Gonzalez, Raul},
  journal={Nature human behaviour},
  volume={3},
  number={7},
  pages={692--701},
  year={2019},
  publisher={Nature Publishing Group UK London}
}

@article{lehtonen2018bilingualism,
  title={Is bilingualism associated with enhanced executive functioning in adults? A meta-analytic review.},
  author={Lehtonen, Minna and Soveri, Anna and Laine, Aini and J{\"a}rvenp{\"a}{\"a}, Janica and De Bruin, Angela and Antfolk, Jan},
  journal={Psychological bulletin},
  volume={144},
  number={4},
  pages={394},
  year={2018},
  publisher={American psychological association}
}

@article{McInnes2018, doi = {10.21105/joss.00861}, url = {https://doi.org/10.21105/joss.00861}, year = {2018}, publisher = {The Open Journal}, volume = {3}, number = {29}, pages = {861}, author = {McInnes, Leland and Healy, John and Saul, Nathaniel and Großberger, Lukas}, title = {UMAP: Uniform Manifold Approximation and Projection}, journal = {Journal of Open Source Software} }

@inproceedings{feng-etal-2024-child,
    title = "Is Child-Directed Speech Effective Training Data for Language Models?",
    author = "Feng, Steven Y.  and
      Goodman, Noah D.  and
      Frank, Michael C.",
    editor = "Al-Onaizan, Yaser  and
      Bansal, Mohit  and
      Chen, Yun-Nung",
    booktitle = "Proceedings of the 2024 Conference on Empirical Methods in Natural Language Processing",
    month = nov,
    year = "2024",
    address = "Miami, Florida, USA",
    publisher = "Association for Computational Linguistics",
    url = "https://aclanthology.org/2024.emnlp-main.1231/",
    doi = "10.18653/v1/2024.emnlp-main.1231",
    pages = "22055--22071"
}

@inproceedings{kallini-etal-2024-mission,
    title = "Mission: Impossible Language Models",
    author = "Kallini, Julie  and
      Papadimitriou, Isabel  and
      Futrell, Richard  and
      Mahowald, Kyle  and
      Potts, Christopher",
    editor = "Ku, Lun-Wei  and
      Martins, Andre  and
      Srikumar, Vivek",
    booktitle = "Proceedings of the 62nd Annual Meeting of the Association for Computational Linguistics (Volume 1: Long Papers)",
    month = aug,
    year = "2024",
    address = "Bangkok, Thailand",
    publisher = "Association for Computational Linguistics",
    url = "https://aclanthology.org/2024.acl-long.787/",
    doi = "10.18653/v1/2024.acl-long.787",
    pages = "14691--14714"
}

@article{constantinescu-etal-2025-investigating,
    title = "Investigating Critical Period Effects in Language Acquisition through Neural Language Models",
    author = "Constantinescu, Ionut  and
      Pimentel, Tiago  and
      Cotterell, Ryan  and
      Warstadt, Alex",
    journal = "Transactions of the Association for Computational Linguistics",
    volume = "13",
    year = "2025",
    address = "Cambridge, MA",
    publisher = "MIT Press",
    url = "https://aclanthology.org/2025.tacl-1.5/",
    doi = "10.1162/tacl_a_00725",
    pages = "96--120"
}

@article{jumelet2025babybabellm,
  title={BabyBabelLM: A Multilingual Benchmark of Developmentally Plausible Training Data},
  author={Jumelet, Jaap and Fourtassi, Abdellah and Haga, Akari and Bunzeck, Bastian and Shandilya, Bhargav and Galvan-Sosa, Diana and Haznitrama, Faiz Ghifari and Padovani, Francesca and Meyer, Francois and Hu, Hai and others},
  journal={arXiv preprint arXiv:2510.10159},
  year={2025}
}

@article{feng2026baby,
  title={Baby Scale: Investigating Models Trained on Individual Children's Language Input},
  author={Feng, Steven Y and Tan, Alvin WM and Frank, Michael C},
  journal={arXiv preprint arXiv:2603.29522},
  year={2026}
}

@article{kiss-strunk-2006-unsupervised,
    title = "Unsupervised Multilingual Sentence Boundary Detection",
    author = "Kiss, Tibor  and
      Strunk, Jan",
    journal = "Computational Linguistics",
    volume = "32",
    number = "4",
    year = "2006",
    url = "https://aclanthology.org/J06-4003/",
    doi = "10.1162/coli.2006.32.4.485",
    pages = "485--525"
}

@inproceedings{bird-2006-nltk,
    title = "{NLTK}: The {N}atural {L}anguage {T}oolkit",
    author = "Bird, Steven",
    editor = "Curran, James",
    booktitle = "Proceedings of the {COLING}/{ACL} 2006 Interactive Presentation Sessions",
    month = jul,
    year = "2006",
    address = "Sydney, Australia",
    publisher = "Association for Computational Linguistics",
    url = "https://aclanthology.org/P06-4018/",
    doi = "10.3115/1225403.1225421",
    pages = "69--72"
}

@inproceedings{mosbach-etal-2020-interplay-fine,
    title = "{O}n the {I}nterplay {B}etween {F}ine-tuning and {S}entence-level {P}robing for {L}inguistic {K}nowledge in {P}re-trained {T}ransformers",
    author = "Mosbach, Marius  and
      Khokhlova, Anna  and
      Hedderich, Michael A.  and
      Klakow, Dietrich",
    editor = "Cohn, Trevor  and
      He, Yulan  and
      Liu, Yang",
    booktitle = "Findings of the Association for Computational Linguistics: EMNLP 2020",
    month = nov,
    year = "2020",
    address = "Online",
    publisher = "Association for Computational Linguistics",
    url = "https://aclanthology.org/2020.findings-emnlp.227/",
    doi = "10.18653/v1/2020.findings-emnlp.227",
    pages = "2502--2516"
}

@article{brinchmann2019there,
  title={Is there a direct relation between the development of vocabulary and grammar?},
  author={Brinchmann, Ellen Ir{\'e}n and Braeken, Johan and Lyster, Solveig-Alma Halaas},
  journal={Developmental science},
  volume={22},
  number={1},
  pages={e12709},
  year={2019},
  publisher={Wiley Online Library}
}

@article{chen2026assessing,
  title={Assessing the Impact of Language Disparity on Multilingual Linguistic Ability in Large Language Models},
  author={Chen, Zhanyu and Jumelet, Jaap},
  journal={arXiv preprint arXiv:2610.00540},
  year={2026}
}

@article{kuwanto2024linguistics,
  title={Linguistics theory meets llm: Code-switched text generation via equivalence constrained large language models},
  author={Kuwanto, Garry and Agarwal, Chaitanya and Winata, Genta Indra and Wijaya, Derry Tanti},
  journal={arXiv preprint arXiv:2410.22660},
  year={2024}
}

@inproceedings{zeng2024leveraging,
  title={Leveraging large language models for code-mixed data augmentation in sentiment analysis},
  author={Zeng, Linda},
  booktitle={Proceedings of the Second Workshop on Social Influence in Conversations (SICon 2024)},
  pages={85--101},
  year={2024}
}

@article{xie2025switchlingua,
  title={SwitchLingua: The First Large-Scale Multilingual and Multi-Ethnic Code-Switching Dataset},
  author={Xie, Peng and Liu, Xingyuan and Chan, Tsz Wai and Bie, Yequan and Song, Yangqiu and Wang, Yang and Chen, Hao and Chen, Kani},
  journal={arXiv preprint arXiv:2506.00087},
  year={2025}
}

@inproceedings{chang-etal-2024-multilinguality,
    title = "When Is Multilinguality a Curse? Language Modeling for 250 High- and Low-Resource Languages",
    author = "Chang, Tyler A.  and
      Arnett, Catherine  and
      Tu, Zhuowen  and
      Bergen, Benjamin K.",
    editor = "Al-Onaizan, Yaser  and
      Bansal, Mohit  and
      Chen, Yun-Nung",
    booktitle = "Proceedings of the 2024 Conference on Empirical Methods in Natural Language Processing",
    month = nov,
    year = "2024",
    address = "Miami, Florida, USA",
    publisher = "Association for Computational Linguistics",
    url = "https://aclanthology.org/2024.emnlp-main.236/",
    doi = "10.18653/v1/2024.emnlp-main.236",
    pages = "4074--4096"
}

@inproceedings{conneau-etal-2020-unsupervised,
    title = "Unsupervised Cross-lingual Representation Learning at Scale",
    author = "Conneau, Alexis  and
      Khandelwal, Kartikay  and
      Goyal, Naman  and
      Chaudhary, Vishrav  and
      Wenzek, Guillaume  and
      Guzm{\'a}n, Francisco  and
      Grave, Edouard  and
      Ott, Myle  and
      Zettlemoyer, Luke  and
      Stoyanov, Veselin",
    editor = "Jurafsky, Dan  and
      Chai, Joyce  and
      Schluter, Natalie  and
      Tetreault, Joel",
    booktitle = "Proceedings of the 58th Annual Meeting of the Association for Computational Linguistics",
    month = jul,
    year = "2020",
    address = "Online",
    publisher = "Association for Computational Linguistics",
    url = "https://aclanthology.org/2020.acl-main.747/",
    doi = "10.18653/v1/2020.acl-main.747",
    pages = "8440--8451"
}

@article{Kuperman2012AgeofacquisitionRF,
  title={Age-of-acquisition ratings for 30,000 English words},
  author={Victor Kuperman and Hans Stadthagen-Gonz{\'a}lez and Marc Brysbaert},
  journal={Behavior Research Methods},
  year={2012},
  volume={44},
  pages={978-990},
  url={https://api.semanticscholar.org/CorpusID:22137152}
}

@inproceedings{wang-etal-2020-negative,
    title = "On Negative Interference in Multilingual Models: Findings and A Meta-Learning Treatment",
    author = "Wang, Zirui  and
      Lipton, Zachary C.  and
      Tsvetkov, Yulia",
    editor = "Webber, Bonnie  and
      Cohn, Trevor  and
      He, Yulan  and
      Liu, Yang",
    booktitle = "Proceedings of the 2020 Conference on Empirical Methods in Natural Language Processing (EMNLP)",
    month = nov,
    year = "2020",
    address = "Online",
    publisher = "Association for Computational Linguistics",
    url = "https://aclanthology.org/2020.emnlp-main.359/",
    doi = "10.18653/v1/2020.emnlp-main.359",
    pages = "4438--4450"
}

@inproceedings{rust-etal-2021-good,
    title = "How Good is Your Tokenizer? On the Monolingual Performance of Multilingual Language Models",
    author = "Rust, Phillip  and
      Pfeiffer, Jonas  and
      Vuli{\'c}, Ivan  and
      Ruder, Sebastian  and
      Gurevych, Iryna",
    editor = "Zong, Chengqing  and
      Xia, Fei  and
      Li, Wenjie  and
      Navigli, Roberto",
    booktitle = "Proceedings of the 59th Annual Meeting of the Association for Computational Linguistics and the 11th International Joint Conference on Natural Language Processing (Volume 1: Long Papers)",
    month = aug,
    year = "2021",
    address = "Online",
    publisher = "Association for Computational Linguistics",
    url = "https://aclanthology.org/2021.acl-long.243/",
    doi = "10.18653/v1/2021.acl-long.243",
    pages = "3118--3135"
}

@article{zhuang2023visual,
  title={Visual Grounding Helps Learn Word Meanings in Low-Data Regimes},
  author={Zhuang, Chengxu and Fedorenko, Evelina and Andreas, Jacob},
  journal={arXiv preprint arXiv:2310.13257},
  year={2023}
}

@article{wordsim-353,
author = {Finkelstein, Lev and Gabrilovich, Evgeniy and Matias, Yossi and Rivlin, Ehud and Solan, Zach and Wolfman, Gadi and Ruppin, Eytan},
year = {2001},
month = {01},
pages = {406-414},
title = {Placing search in context: The concept revisited},
volume = {20},
journal = {ACM Transactions on Information Systems - TOIS},
doi = {10.1145/503104.503110}
}

@article{hill-etal-2015-simlex,
    title = "{S}im{L}ex-999: Evaluating Semantic Models With (Genuine) Similarity Estimation",
    author = "Hill, Felix  and
      Reichart, Roi  and
      Korhonen, Anna",
    journal = "Computational Linguistics",
    volume = "41",
    number = "4",
    month = dec,
    year = "2015",
    address = "Cambridge, MA",
    publisher = "MIT Press",
    url = "https://aclanthology.org/J15-4004",
    doi = "10.1162/COLI_a_00237",
    pages = "665--695",
}

@inproceedings{gerz-etal-2016-simverb,
    title = "{S}im{V}erb-3500: A Large-Scale Evaluation Set of Verb Similarity",
    author = "Gerz, Daniela  and
      Vuli{\'c}, Ivan  and
      Hill, Felix  and
      Reichart, Roi  and
      Korhonen, Anna",
    editor = "Su, Jian  and
      Duh, Kevin  and
      Carreras, Xavier",
    booktitle = "Proceedings of the 2016 Conference on Empirical Methods in Natural Language Processing",
    month = nov,
    year = "2016",
    address = "Austin, Texas",
    publisher = "Association for Computational Linguistics",
    url = "https://aclanthology.org/D16-1235",
    doi = "10.18653/v1/D16-1235",
    pages = "2173--2182",
}

@inproceedings{bruni-etal-2012-distributional,
    title = "Distributional Semantics in Technicolor",
    author = "Bruni, Elia  and
      Boleda, Gemma  and
      Baroni, Marco  and
      Tran, Nam-Khanh",
    editor = "Li, Haizhou  and
      Lin, Chin-Yew  and
      Osborne, Miles  and
      Lee, Gary Geunbae  and
      Park, Jong C.",
    booktitle = "Proceedings of the 50th Annual Meeting of the Association for Computational Linguistics (Volume 1: Long Papers)",
    month = jul,
    year = "2012",
    address = "Jeju Island, Korea",
    publisher = "Association for Computational Linguistics",
    url = "https://aclanthology.org/P12-1015",
    pages = "136--145",
}

@inproceedings{rg-65,
    title = "Contextual Correlates of Synonymy",
    author = "Rubenstein, Herbert and Goodenough, John B.",
    year = 1965,
    booktitle = "Communications of the ACM (CACM) 8 (10)"}

@inproceedings{wu-dredze-2020-languages,
    title = "Are All Languages Created Equal in Multilingual {BERT}?",
    author = "Wu, Shijie  and
      Dredze, Mark",
    editor = "Gella, Spandana  and
      Welbl, Johannes  and
      Rei, Marek  and
      Petroni, Fabio  and
      Lewis, Patrick  and
      Strubell, Emma  and
      Seo, Minjoon  and
      Hajishirzi, Hannaneh",
    booktitle = "Proceedings of the 5th Workshop on Representation Learning for NLP",
    month = jul,
    year = "2020",
    address = "Online",
    publisher = "Association for Computational Linguistics",
    url = "https://aclanthology.org/2020.repl4nlp-1.16/",
    doi = "10.18653/v1/2020.repl4nlp-1.16",
    pages = "120--130"
}

@inproceedings{fujinuma-etal-2022-match,
    title = "Match the Script, Adapt if Multilingual: Analyzing the Effect of Multilingual Pretraining on Cross-lingual Transferability",
    author = "Fujinuma, Yoshinari  and
      Boyd-Graber, Jordan  and
      Kann, Katharina",
    editor = "Muresan, Smaranda  and
      Nakov, Preslav  and
      Villavicencio, Aline",
    booktitle = "Proceedings of the 60th Annual Meeting of the Association for Computational Linguistics (Volume 1: Long Papers)",
    month = may,
    year = "2022",
    address = "Dublin, Ireland",
    publisher = "Association for Computational Linguistics",
    url = "https://aclanthology.org/2022.acl-long.106/",
    doi = "10.18653/v1/2022.acl-long.106",
    pages = "1500--1512"
}

@inproceedings{pyysalo-etal-2021-wikibert,
    title = "{W}iki{BERT} Models: Deep Transfer Learning for Many Languages",
    author = "Pyysalo, Sampo  and
      Kanerva, Jenna  and
      Virtanen, Antti  and
      Ginter, Filip",
    editor = "Dobnik, Simon  and
      {\O}vrelid, Lilja",
    booktitle = "Proceedings of the 23rd Nordic Conference on Computational Linguistics (NoDaLiDa)",
    month = may # " 31--2 " # jun,
    year = "2021",
    address = "Reykjavik, Iceland (Online)",
    publisher = {Link{\"o}ping University Electronic Press, Sweden},
    url = "https://aclanthology.org/2021.nodalida-main.1/",
    pages = "1--10"
}

@article{ruder2017overview,
  title={An overview of multi-task learning in deep neural networks},
  author={Ruder, Sebastian},
  journal={arXiv preprint arXiv:1706.05098},
  year={2017}
}

@article{li2004early,
  title={Early lexical development in a self-organizing neural network},
  author={Li, Ping and Farkas, Igor and MacWhinney, Brian},
  journal={Neural networks},
  volume={17},
  number={8-9},
  pages={1345--1362},
  year={2004},
  publisher={Elsevier}
}

@article{kaushanskaya2023combining,
  title={Combining languages in bilingual input: Using experimental evidence to formulate bilingual exposure strategies},
  author={Kaushanskaya, Margarita},
  journal={Journal of Speech, Language, and Hearing Research},
  volume={66},
  number={12},
  pages={4771--4784},
  year={2023},
  publisher={American Speech-Language-Hearing Association}
}

@article{libersky2024effects,
  title={Effects of dual-and single-language exposure on children’s word learning: Experimentally testing the role of competition},
  author={Libersky, Emma and Slawny, Caitlyn and Kaushanskaya, Margarita},
  journal={Journal of experimental child psychology},
  volume={244},
  pages={105953},
  year={2024},
  publisher={Elsevier}
}

@article{potter2025infants,
  title={Infants Do Not Reliably Track When Bilingual Speakers Switch Languages},
  author={Potter, Christine E and Lew-Williams, Casey},
  journal={Behavioral Sciences},
  volume={15},
  number={10},
  pages={1427},
  year={2025},
  publisher={MDPI}
}

@article{prior2007translation,
  title={Translation norms for English and Spanish: The role of lexical variables, word class, and L2 proficiency in negotiating translation ambiguity},
  author={Prior, Anat and MacWhinney, Brian and Kroll, Judith F},
  journal={Behavior Research Methods},
  volume={39},
  number={4},
  pages={1029--1038},
  year={2007},
  publisher={Springer}
}

@article{byers2013bilingualism,
  title={Bilingualism in the early years: What the science says},
  author={Byers-Heinlein, Krista and Lew-Williams, Casey},
  journal={LEARNing landscapes},
  volume={7},
  number={1},
  pages={95},
  year={2013}
}

@article{byers2010roots,
  title={The roots of bilingualism in newborns},
  author={Byers-Heinlein, Krista and Burns, Tracey C and Werker, Janet F},
  journal={Psychological science},
  volume={21},
  number={3},
  pages={343--348},
  year={2010},
  publisher={Sage Publications Sage CA: Los Angeles, CA}
}

@article{mehler1988precursor,
  title={A precursor of language acquisition in young infants},
  author={Mehler, Jacques and Jusczyk, Peter and Lambertz, Ghislaine and Halsted, Nilofar and Bertoncini, Josiane and Amiel-Tison, Claudine},
  journal={Cognition},
  volume={29},
  number={2},
  pages={143--178},
  year={1988},
  publisher={Elsevier}
}

@book{ronjat1913developpement,
  title={Le d{\'e}veloppement du langage observ{\'e} chez un enfant bilingue},
  author={Ronjat, Jules},
  year={1913},
  publisher={Champion}
}

@article{de2007parental,
  title={Parental language input patterns and children's bilingual use},
  author={De Houwer, Annick},
  journal={Applied psycholinguistics},
  volume={28},
  number={3},
  pages={411--424},
  year={2007},
  publisher={Cambridge University Press}
}

@article{kovacs2009early,
  title={Early bilingualism enhances mechanisms of false-belief reasoning},
  author={Kov{\'a}cs, {\'A}gnes Melinda},
  journal={Developmental science},
  volume={12},
  number={1},
  pages={48--54},
  year={2009},
  publisher={Wiley Online Library}
}

@article{brito2012influence,
  title={Influence of bilingualism on memory generalization during infancy},
  author={Brito, Natalie and Barr, Rachel},
  journal={Developmental science},
  volume={15},
  number={6},
  pages={812--816},
  year={2012},
  publisher={Wiley Online Library}
}

@article{kovacs2009cognitive,
  title={Cognitive gains in 7-month-old bilingual infants},
  author={Kov{\'a}cs, {\'A}gnes Melinda and Mehler, Jacques},
  journal={Proceedings of the National Academy of Sciences},
  volume={106},
  number={16},
  pages={6556--6560},
  year={2009},
  publisher={National Academy of Sciences}
}

@article{poulin2011effects,
  title={The effects of bilingualism on toddlers’ executive functioning},
  author={Poulin-Dubois, Diane and Blaye, Agnes and Coutya, Julie and Bialystok, Ellen},
  journal={Journal of experimental child psychology},
  volume={108},
  number={3},
  pages={567--579},
  year={2011},
  publisher={Elsevier}
}

@article{bialystok2012bilingualism,
  title={Bilingualism: Consequences for mind and brain},
  author={Bialystok, Ellen and Craik, Fergus IM and Luk, Gigi},
  journal={Trends in cognitive sciences},
  volume={16},
  number={4},
  pages={240--250},
  year={2012},
  publisher={Elsevier}
}

@article{goetz2003effects,
  title={The effects of bilingualism on theory of mind development},
  author={Goetz, Peggy J},
  journal={Bilingualism: Language and cognition},
  volume={6},
  number={1},
  pages={1--15},
  year={2003},
  publisher={Cambridge University Press}
}

@inproceedings{oba-etal-2023-second,
    title = "Second Language Acquisition of Neural Language Models",
    author = "Oba, Miyu  and
      Kuribayashi, Tatsuki  and
      Ouchi, Hiroki  and
      Watanabe, Taro",
    editor = "Rogers, Anna  and
      Boyd-Graber, Jordan  and
      Okazaki, Naoaki",
    booktitle = "Findings of the Association for Computational Linguistics: ACL 2023",
    month = jul,
    year = "2023",
    address = "Toronto, Canada",
    publisher = "Association for Computational Linguistics",
    url = "https://aclanthology.org/2023.findings-acl.856/",
    doi = "10.18653/v1/2023.findings-acl.856",
    pages = "13557--13572"
}

@article{poplack1980codeswitching,
  author  = {Poplack, Shana},
  title   = {Sometimes I’ll Start a Sentence in Spanish y Termino en Español: Toward a Typology of Code-Switching},
  journal = {Linguistics},
  volume  = {18},
  number  = {7-8},
  pages   = {581--618},
  year    = {1980},
  doi     = {10.1515/ling.1980.18.7-8.581}
}

@article{bosch1997native,
  title={Native-language recognition abilities in 4-month-old infants from monolingual and bilingual environments},
  author={Bosch, Laura and Sebasti{\'a}n-Gall{\'e}s, N{\'u}ria},
  journal={Cognition},
  volume={65},
  number={1},
  pages={33--69},
  year={1997},
  publisher={Elsevier}
}

@article{nazzi2000language,
  title={Language discrimination by English-learning 5-month-olds: Effects of rhythm and familiarity},
  author={Nazzi, Thierry and Jusczyk, Peter W and Johnson, Elizabeth K},
  journal={Journal of Memory and Language},
  volume={43},
  number={1},
  pages={1--19},
  year={2000},
  publisher={Elsevier}
}

@inproceedings{philippy-etal-2023-towards,
    title = "Towards a Common Understanding of Contributing Factors for Cross-Lingual Transfer in Multilingual Language Models: A Review",
    author = "Philippy, Fred  and
      Guo, Siwen  and
      Haddadan, Shohreh",
    editor = "Rogers, Anna  and
      Boyd-Graber, Jordan  and
      Okazaki, Naoaki",
    booktitle = "Proceedings of the 61st Annual Meeting of the Association for Computational Linguistics (Volume 1: Long Papers)",
    month = jul,
    year = "2023",
    address = "Toronto, Canada",
    publisher = "Association for Computational Linguistics",
    url = "https://aclanthology.org/2023.acl-long.323/",
    doi = "10.18653/v1/2023.acl-long.323",
    pages = "5877--5891"
}

@article{10.1145/3727339,
author = {Doddapaneni, Sumanth and Ramesh, Gowtham and Khapra, Mitesh and Kunchukuttan, Anoop and Kumar, Pratyush},
title = {A Primer on Pretrained Multilingual Language Models},
year = {2025},
issue_date = {September 2025},
publisher = {Association for Computing Machinery},
address = {New York, NY, USA},
volume = {57},
number = {9},
issn = {0360-0300},
url = {https://doi.org/10.1145/3727339},
doi = {10.1145/3727339},
journal = {ACM Comput. Surv.},
month = may,
articleno = {232},
numpages = {39}
}

@inproceedings{wu2019beto,
  title={Beto, bentz, becas: The surprising cross-lingual effectiveness of BERT},
  author={Wu, Shijie and Dredze, Mark},
  booktitle={Proceedings of the 2019 conference on empirical methods in natural language processing and the 9th international joint conference on natural language processing (EMNLP-IJCNLP)},
  pages={833--844},
  year={2019}
}

@inproceedings{misra-mahowald-2024-language,
    title = "Language Models Learn Rare Phenomena from Less Rare Phenomena: The Case of the Missing {AANN}s",
    author = "Misra, Kanishka  and
      Mahowald, Kyle",
    editor = "Al-Onaizan, Yaser  and
      Bansal, Mohit  and
      Chen, Yun-Nung",
    booktitle = "Proceedings of the 2024 Conference on Empirical Methods in Natural Language Processing",
    month = nov,
    year = "2024",
    address = "Miami, Florida, USA",
    publisher = "Association for Computational Linguistics",
    url = "https://aclanthology.org/2024.emnlp-main.53/",
    doi = "10.18653/v1/2024.emnlp-main.53",
    pages = "913--929"
}

@article{kaushanskaya2009bilingual,
  title={The bilingual advantage in novel word learning},
  author={Kaushanskaya, Margarita and Marian, Viorica},
  journal={Psychonomic bulletin \& review},
  volume={16},
  number={4},
  pages={705--710},
  year={2009},
  publisher={Springer}
}

@inproceedings{wang-etal-2025-investigating-scaling,
    title = "Investigating and Scaling up Code-Switching for Multilingual Language Model Pre-Training",
    author = "Wang, Zhijun  and
      Li, Jiahuan  and
      Zhou, Hao  and
      Weng, Rongxiang  and
      Wang, Jingang  and
      Huang, Xin  and
      Han, Xue  and
      Feng, Junlan  and
      Deng, Chao  and
      Huang, Shujian",
    editor = "Che, Wanxiang  and
      Nabende, Joyce  and
      Shutova, Ekaterina  and
      Pilehvar, Mohammad Taher",
    booktitle = "Findings of the Association for Computational Linguistics: ACL 2025",
    month = jul,
    year = "2025",
    address = "Vienna, Austria",
    publisher = "Association for Computational Linguistics",
    url = "https://aclanthology.org/2025.findings-acl.575/",
    doi = "10.18653/v1/2025.findings-acl.575",
    pages = "11032--11046",
    ISBN = "979-8-89176-256-5"
}

@article{radford2019language,
  title={Language models are unsupervised multitask learners},
  author={Radford, Alec and Wu, Jeffrey and Child, Rewon and Luan, David and Amodei, Dario and Sutskever, Ilya and others},
  journal={OpenAI blog},
  volume={1},
  number={8},
  pages={9},
  year={2019}
}

@inproceedings{yoo-etal-2025-code-switching,
    title = "Code-Switching Curriculum Learning for Multilingual Transfer in {LLM}s",
    author = "Yoo, Haneul  and
      Park, Cheonbok  and
      Yun, Sangdoo  and
      Oh, Alice  and
      Lee, Hwaran",
    editor = "Che, Wanxiang  and
      Nabende, Joyce  and
      Shutova, Ekaterina  and
      Pilehvar, Mohammad Taher",
    booktitle = "Findings of the Association for Computational Linguistics: ACL 2025",
    month = jul,
    year = "2025",
    address = "Vienna, Austria",
    publisher = "Association for Computational Linguistics",
    url = "https://aclanthology.org/2025.findings-acl.407/",
    doi = "10.18653/v1/2025.findings-acl.407",
    pages = "7816--7836",
    ISBN = "979-8-89176-256-5"
}

@inproceedings{ibanez2010inhibitory,
  title={Inhibitory control in 8-month-old monolingual and bilingual infants: Evidence from an anticipatory eye movement task},
  author={Ibanez-Lillo, Alexandra and Pons, Ferran and Costa, Albert and Sebastian-Galles, Nunia},
  booktitle={Poster presented at the 22nd Biennial International Conference on Infant Studies, Baltimore, MD},
  year={2010}
}

@article{carlson2008bilingual,
  title={Bilingual experience and executive functioning in young children},
  author={Carlson, Stephanie M and Meltzoff, Andrew N},
  journal={Developmental science},
  volume={11},
  number={2},
  pages={282--298},
  year={2008},
  publisher={Wiley Online Library}
}

@article{kapantzoglou2021code,
  title={Code-switching and language proficiency in bilingual children with and without developmental language disorder},
  author={Kapantzoglou, Maria and Brown, Julie Esparza and Cycyk, Lauren M and Fergadiotis, Gerasimos},
  journal={Journal of Speech, Language, and Hearing Research},
  volume={64},
  number={5},
  pages={1605--1620},
  year={2021},
  publisher={American Speech-Language-Hearing Association}
}

@article{hoff2012dual,
  title={Dual language exposure and early bilingual development},
  author={Hoff, Erika and Core, Cynthia and Place, Silvia and Rumiche, Rosario and Se{\~n}or, Melissa and Parra, Marisol},
  journal={Journal of child language},
  volume={39},
  number={1},
  pages={1--27},
  year={2012},
  publisher={Cambridge University Press}
}

@article{comeau2003modeling,
  title={The modeling hypothesis and child bilingual codemixing},
  author={Comeau, Liane and Genesee, Fred and Lapaquette, Lindsay},
  journal={International journal of Bilingualism},
  volume={7},
  number={2},
  pages={113--126},
  year={2003},
  publisher={SAGE Publications Sage UK: London, England}
}

@book{pearson2008raising,
  title={Raising a bilingual child},
  author={Pearson, Barbara Zurer},
  year={2008},
  publisher={Random House Reference}
}

@article{eldan2023tinystories,
  title={Tinystories: How small can language models be and still speak coherent english?},
  author={Eldan, Ronen and Li, Yuanzhi},
  journal={arXiv preprint arXiv:2305.07759},
  year={2023}
}

@article{frank2023openly,
  title={Openly accessible LLMs can help us to understand human cognition},
  author={Frank, Michael C},
  journal={Nature Human Behaviour},
  volume={7},
  number={11},
  pages={1825--1827},
  year={2023},
  publisher={Nature Publishing Group UK London}
}

@inproceedings{zhang-etal-2021-need,
    title = "When Do You Need Billions of Words of Pretraining Data?",
    author = "Zhang, Yian  and
      Warstadt, Alex  and
      Li, Xiaocheng  and
      Bowman, Samuel R.",
    editor = "Zong, Chengqing  and
      Xia, Fei  and
      Li, Wenjie  and
      Navigli, Roberto",
    booktitle = "Proceedings of the 59th Annual Meeting of the Association for Computational Linguistics and the 11th International Joint Conference on Natural Language Processing (Volume 1: Long Papers)",
    month = aug,
    year = "2021",
    address = "Online",
    publisher = "Association for Computational Linguistics",
    url = "https://aclanthology.org/2021.acl-long.90/",
    doi = "10.18653/v1/2021.acl-long.90",
    pages = "1112--1125"
}

@inproceedings{huebner-etal-2021-babyberta,
    title = "{B}aby{BERT}a: Learning More Grammar With Small-Scale Child-Directed Language",
    author = "Huebner, Philip A.  and
      Sulem, Elior  and
      Cynthia, Fisher  and
      Roth, Dan",
    editor = "Bisazza, Arianna  and
      Abend, Omri",
    booktitle = "Proceedings of the 25th Conference on Computational Natural Language Learning",
    month = nov,
    year = "2021",
    address = "Online",
    publisher = "Association for Computational Linguistics",
    url = "https://aclanthology.org/2021.conll-1.49/",
    doi = "10.18653/v1/2021.conll-1.49",
    pages = "624--646"
}

@inproceedings{warstadt-etal-2023-findings,
    title = "Findings of the {B}aby{LM} Challenge: Sample-Efficient Pretraining on Developmentally Plausible Corpora",
    author = "Warstadt, Alex  and
      Mueller, Aaron  and
      Choshen, Leshem  and
      Wilcox, Ethan  and
      Zhuang, Chengxu  and
      Ciro, Juan  and
      Mosquera, Rafael  and
      Paranjabe, Bhargavi  and
      Williams, Adina  and
      Linzen, Tal  and
      Cotterell, Ryan",
    editor = "Warstadt, Alex  and
      Mueller, Aaron  and
      Choshen, Leshem  and
      Wilcox, Ethan  and
      Zhuang, Chengxu  and
      Ciro, Juan  and
      Mosquera, Rafael  and
      Paranjabe, Bhargavi  and
      Williams, Adina  and
      Linzen, Tal  and
      Cotterell, Ryan",
    booktitle = "Proceedings of the BabyLM Challenge at the 27th Conference on Computational Natural Language Learning",
    month = dec,
    year = "2023",
    address = "Singapore",
    publisher = "Association for Computational Linguistics",
    url = "https://aclanthology.org/2023.conll-babylm.1/",
    doi = "10.18653/v1/2023.conll-babylm.1",
    pages = "1--34"
}

\appendix

\newpage

\section{Additional Methods}

\subsection{Data Collection Details and Examples}\label{appendix:datacollection}

We provide additional details on the data collection process for the four parallel 100M-word datasets -- English, L2, sentence-level code-switching, and word-level code-switching -- repeated for three L2s. We present an example of the same dialogue across these language conditions in Spanish in Table~\ref{tab:examples}. Dialogues were seeded with age-specific vocabulary lists and varied in length, topic, and interaction type (e.g., storytelling, routines, questioning, emotional exchanges). The specific models used for data generation were \textit{gpt-4o-mini-2024-07-18}, which is the 8-billion parameter version of GPT-4o with training data up to July 2024, for English, Spanish, and word-level code-switching data, and \textit{gemini-3.1-flash-lite} (training data up to January 2025) for Chinese and Dutch data. We used a seed of 42 for all data collection methods.

To generate the English data, we constructed dialogues across four categories defined by speaker (mother vs. father) and context (home vs. public). We chose this distinction to explore the key properties of bilingual environments, where language use varies based on both interlocutor and interactional setting.  Accordingly, we generated four balanced subsets: mother–home (25M tokens), father–home (25M), mother–public (25M), and father–public (25M).

Following \cite{feng-etal-2024-child}, we introduced additional controls to increase the diversity of the generated conversations. For each GPT-4 prompt, we specified the child’s age (2, 5, 10, or 15), the conversation type (Table \ref{tab:TD_conversation_types}), and the approximate dialogue length (5 or 10 turns). To further diversify lexical content, each prompt included a constraint requiring the use of three target words—one noun, one verb, and one adjective—sampled from Wordbank CDI \citep{frank2023openly} for younger ages (2, 5) and from Age of Acquisition norms \citep{Kuperman2012AgeofacquisitionRF} for older ages (10, 15). The selection of target words were sorted based on context (home vs. public), while the speaker (mother vs. father) determined the speaker labels used in the dialogue. We additionally collect metadata describing the participants and the conversational context or setting. Only the dialogue portions were retained for downstream experiments. We did not use in-context prompts.

The prompt used to generate the original English conversations is provided below (max tokens = 2000, temperature = 0.5).

\begin{quote}
\small
\texttt{Please construct a realistic, approximately \{5, 10\}-turn dialogue directly involving a \{2, 5, 10, 15\}-year-old \{toddler, child, teenager\} as a participant. The \{toddler, child, teenager\} is the central participant, and most or all speech should be directed toward them.
Use only vocabulary that a typical \{2, 5, 10, 15\}-year-old \{toddler, child, teenager\} would understand.
The parent participant and the setting are fixed for this dialogue and must be: **\{Mom, Dad\}** in a \{home, public\}\footnote{In all instances, "public" is written as "public (school or community)".} environment.
The dialogue must take place entirely in a \{home, public\} environment, with physical, social, or sensory details that make it impossible to mistake for a \{home, public\}\footnote{The alternative environment (i.e., the one not selected for this instance) is referenced here. For example, if the dialogue is set in a home environment, "public" is used in this clause, and vice versa.} environment.
The topic does not need to be about \{home, public\}, but all lines should naturally reference people, activities, or surroundings specific to this setting. Include at least two clear details that could not occur in a \{public, home\} environment.
The dialogue should be \{type\}\footnote{A random conversation type along with its explanation is sampled each time from the ones in Table \ref{tab:TD_conversation_types}.} and must include the verb '\{verb\}', the noun '\{noun\}', and the adjective '\{adjective\}'.
Strictly do not use or invent any proper names for any participant or in the dialogue. Use only the labels **\{Mom, Dad\}** and **Child** when referring to participants.
Ensure that at least \{5, 10\} turns of direct speech are included under the `DIALOGUE:' section (no summaries or narration in place of dialogue). Participant labels must be surrounded by double asterisks, e.g., '**participant**'.
Output must strictly follow this format:
PARTICIPANTS:  
[List and describe only **\{Mom, Dad\}** and **Child** without names]
SETTING:  
[Briefly describe the context or setting]
DIALOGUE:  
[List \{turn\} turns of dialogue separated by \textbackslash n\textbackslash n]
Remember, the dialogue must be realistic and likely to occur in the real world.
}

\end{quote}

To construct L2 versions of the generated dialogues, we prompted GPT-4 to translate all English conversations while preserving their meaning and tone. The prompt used for translation is shown below, replacing \texttt{[TARGET LANGUAGE]} with Spanish, Chinese, or Dutch (max tokens = 2000, temperature = 0.3).

\begin{quote}
\small
\texttt{Translate the following English text (beginning at the word `PARTICIPANTS') into [TARGET LANGUAGE] while preserving its meaning and tone. Do not add explanations or comments. Output only the translated text.
}
\end{quote}

To generate word-level code-switched dialogues, we prompted GPT-4 to rewrite English conversations using English–Spanish code-switching, guided by constraints from \cite{poplack1980codeswitching,kuwanto2024linguistics}. We adapted these prompts for English–Chinese and English–Dutch to account for language-specific grammatical patterns. The English–Spanish prompt is shown below (max tokens = 2000, temperature = 0.5).

\begin{quote}
\small
\texttt{Rewrite the following dialogue using naturalistic Code-switching (word)\footnote{Code-switching (word) code-switching is an equivalent term to refer to word-level code-switching.} English-Spanish code-switching as spoken by bilingual U.S. families. Keep the exact semantic content and line-by-line structure. Switch only within sentences, not between them.
Make switches at natural, grammatically permissible boundaries such as clause edges, prepositional phrases, adverbial phrases, discourse markers, and emotionally marked expressions. Maintain a warm, conversational family tone. Use English for abstract, explanatory, or academic content and Spanish for emotional, relational, culturally grounded, or intimate phrases. Avoid random or token-level switches unless a sentence is already only a single word or idiom.
All code-switching must obey Poplack's constraints:
Free Morpheme Constraint (no switching within bound morphemes; forms like walk-ando or eat-iendo are not allowed),
Equivalence Constraint (switch only where English and Spanish surface structures align; bad examples: El man que came ayer wants John comprar un car nuevo, I think que...; acceptable: Tell Larry que se calle la boca), and
Functional Head Constraint (do not switch between determiners, complementizers, auxiliaries, or other functional heads and their complements; avoid forms like para the kids, está going).
Maintain realistic bilingual rhythm, sociolinguistic authenticity, and avoid textbook Spanish. Ensure there is an even amount of switching across sentences.
Do not add, remove, or alter semantic content. Include PARTICIPANTS and SETTING in English. Only change the language mixture. Keep the dialogue’s line structure exactly as given.
Apply all constraints to this dialogue:
}
\end{quote}

We generated sentence-level code-switched dialogues by aligning each English dialogue with its GPT-4-generated L2 translation. We segmented English, Spanish, and Dutch texts using NLTK sentence tokenization (\texttt{sent\_tokenize}) and Chinese texts using regular expressions. For each aligned sentence pair, we randomly selected either the English or L2 version. Whenever more than three consecutive sentences were selected in the same language, the script selected the other language for the next sentence.

\begin{table*}[t]
\centering
\small
\setlength{\tabcolsep}{6pt}
\begin{tabularx}{\textwidth}{p{0.18\textwidth}X}
\toprule
\textbf{Language} & \textbf{Example} \\
\midrule

English &
**Mom**: Hey, can you please clean up your art supplies before snack time? I don’t want to nag, but it’s getting a bit messy over there.\textbackslash n\textbackslash n
**Child**: But I just started drawing! I want to finish this picture first. It’s going to be a present for Grandma!\textbackslash n\textbackslash n
**Mom**: I know, sweetie, and I love that idea! But if you clean up now, you’ll have extra time to work on it later without any distractions.\textbackslash n\textbackslash n
**Child**: Okay, but can I have a cookie after I clean up? Just one, please!\textbackslash n\textbackslash n
**Mom**: Of course! One cookie after you tidy up. That way, you can enjoy your snack while you finish your drawing. Deal? \\
\midrule

Spanish &
**Mom**: Oye, ¿puedes por favor recoger tus materiales de arte antes de la hora del bocadillo? No quiero ser pesada, pero se está volviendo un poco desordenado allí.\textbackslash n\textbackslash n
**Child**: ¡Pero acabo de empezar a dibujar! Quiero terminar esta imagen primero. ¡Va a ser un regalo para la abuela!\textbackslash n\textbackslash n
**Mom**: Lo sé, cariño, y me encanta esa idea. Pero si limpias ahora, tendrás tiempo extra para trabajar en ello más tarde sin distracciones.\textbackslash n\textbackslash n
**Child**: Está bien, pero ¿puedo comer una galleta después de limpiar? ¡Solo una, por favor!\textbackslash n\textbackslash n
**Mom**: ¡Por supuesto! Una galleta después de que ordenes. Así podrás disfrutar de tu bocadillo mientras terminas tu dibujo. ¿Trato hecho? \\
\midrule

Code-switching (sentence-level) &
**Mom**: Hey, can you please clean up your art supplies before snack time? I don’t want to nag, but it’s getting a bit messy over there.\textbackslash n\textbackslash n
**Child**: ¡Pero acabo de empezar a dibujar! Quiero terminar esta imagen primero. ¡Va a ser un regalo para la abuela!\textbackslash n\textbackslash n
**Mom**: I know, sweetie, and I love that idea! But if you clean up now, you’ll have extra time to work on it later without any distractions.\textbackslash n\textbackslash n
**Child**: Está bien, pero ¿puedo comer una galleta después de limpiar? ¡Solo una, por favor!\textbackslash n\textbackslash n
**Mom**: ¡Por supuesto! One cookie after you tidy up. That way, you can enjoy your snack while you finish your drawing. Deal? \\
\midrule

Code-switching (word-level) &
**Mom**: Hey, ¿puedes please clean up tus materiales de arte before snack time? No quiero nag, but it’s getting un poco messy over there.\textbackslash n\textbackslash n
**Child**: But I just started dibujando! Quiero finish this picture first. It’s going to be un regalo para Grandma!\textbackslash n\textbackslash n
**Mom**: I know, sweetie, y me encanta that idea! But if you limpias now, tendrás extra time to work on it later without distracciones.\textbackslash n\textbackslash n
**Child**: Okay, but ¿puedo have una cookie after I clean up? Just una, please!\textbackslash n\textbackslash n
**Mom**: Of course! Una cookie after you tidy up. Así you can enjoy your snack mientras you finish your drawing. Deal? \\
\bottomrule

\end{tabularx}
\caption{Example of the same generated dialogue (Mom, home, age 10, 5 turns, functional type) across the four language conditions of English-Spanish data collection after preprocessing.}
\label{tab:examples}
\end{table*}

\begin{table*}[t]
\centering
\resizebox{\textwidth}{!}{
\begin{tabular}{c|c}
\toprule
\textbf{Conversation Type} & \textbf{Explanation}\\
\midrule
Explanatory & It should involve explaining something(s) and potentially answering question(s).\\
Functional & It should involve attempting to get something(s) done or accomplishing particular goal(s).\\
Narrative & It should involve telling a story (real or fictional) or sharing/recounting an experience.\\
Argumentative & It should involve conflict(s) or disagreement(s) that lead to an argument.\\
& In most cases, the argument should be resolved, resulting in the \{child, toddler, teenager\} learning.\\
\bottomrule
\end{tabular}
}
\caption{The four conversation types along with their explanations.}
\label{tab:TD_conversation_types}
\end{table*}

Fluency in English, Spanish, and Chinese enabled direct assessment of the generated outputs. Before large-scale generation, prompts were iteratively refined and tested on small batches to assess whether the dialogues exhibited the intended stylistic and linguistic properties. This process focused particularly on the English generation and word-level code-switching prompts, including seed-word selection. Throughout the pipeline, English dialogues, Spanish and Chinese translations, and code-switched outputs were manually spot-checked for linguistic quality and consistency with the assigned speaker roles (Mom or Dad) and language conditions. Additionally, the downstream evaluation results (Section ~\ref{sec:main_expt_results}) themselves provide indirect evidence that the generated data is coherent and learnable, with similar trends to training naturalistic CHILDES data strengthening this finding (Section~\ref{sec:natural}).

\subsection{Data Preprocessing and Statistics}\label{appendix:datapreprocessing}

\begin{table*}[t]
\centering
\small
\begin{tabular}{l|ccc}
\toprule
\textbf{Language} & \textbf{Total Words} & \textbf{Unique (Regex)} & \textbf{Unique (NLTK)}\\
\midrule
English           & 112,350,555 & 31,725 & 45,839 \\
Spanish & 103,559,408 & 79,637 & 111,127\\
Code-switching (sentence-level)  & 102,029,061 & 92,301 & 123,314\\
Code-switching (word-level)  & 109,510,491 & 87,392 & 118,219\\
\bottomrule
\end{tabular}
\caption{Total and unique words for each generated language dataset in English-Spanish after preprocessing and before creating conditions and splitting into train and validation data.}
\label{tab:TD_dataset_stats}
\end{table*}

\begin{table}[t]
\centering
\small
\begin{tabular}{llccc}
\toprule
\textbf{Scale} & \textbf{Condition} & \textbf{Total} & \textbf{Train} & \textbf{Val} \\
\midrule
100M & Full & 799,885 & 759,893 & 39,992 \\
100M & Baseline & 399,949 & 379,953 & 19,996 \\
\midrule
20M & Full & 152,000 & 144,400 & 7,600 \\
20M & Baseline & 76,000 & 72,200 & 3,800 \\
\bottomrule
\end{tabular}
\caption{
Dialogue counts across dataset scales and conditions in English-Spanish after preprocessing, creating conditions, and splitting into training and validation sets. The 100M setting corresponds to the main experiment, while the 20M setting represents the reduced-data condition. Corresponding baselines contain half as many dialogues (and words) as the full datasets. All datasets are split into 95\% training and 5\% validation.
}
\label{tab:dataset_stats}
\end{table}

The parallel English, L2, and code-switching (sentence-level and word-level) dialogue data (generated as described in Appendix \ref{appendix:datacollection}) underwent a uniform preprocessing pipeline. We removed metadata such as speaker descriptions and setting information, replaced typographic quotation marks with ASCII equivalents, and converted raw newline characters into explicit \texttt{\textbackslash n} tokens. Speaker labels were normalized to a fixed set ("Mom", "Dad", and "Child") across all language conditions, including L2, to ensure consistency with code-switched data and avoid the incoherent mixing of speaker labels in different languages within a single dialogue. Each dialogue was terminated with an end-of-sequence token (\texttt{<|endoftext|>}) to ensure compatibility with GPT-2 style language model training.

We used NLTK version 3.9.1 \citep{bird-2006-nltk} for text preprocessing, including the pretrained Punkt sentence tokenizer \citep{kiss-strunk-2006-unsupervised} (\texttt{punkt} and \texttt{punkt\_tab}, downloaded August 11, 2025). We applied \texttt{word\_tokenize} with default settings (\texttt{preserve\_line=False}).

Dataset statistics for four language corpora (in the English-Spanish pair) after preprocessing are reported in Table~\ref{tab:TD_dataset_stats}. Total word counts and whitespace-based vocabulary sizes (Unique (Regex)) were computed using regex-based tokenization, while NLTK vocabulary sizes (Unique (NLTK)) were computed using the Python NLTK package's \texttt{word\_tokenize} function. We observe higher lexical diversity for Spanish and multilingual data than in English. A possible explanation is that when translating from English to Spanish, a single English word can map to multiple Spanish forms because Spanish is a highly inflected language, meaning that it uses more verb conjugations, pronoun forms, articles, and gender/number agreement \citep{prior2007translation}. For example, the English word ``you'' can be translated as "tú," "usted," or "ustedes" depending on context.

After preprocessing, we constructed training conditions by pooling and sampling from the subcorpora described in Appendix~\ref{appendix:datacollection} (each of the four corpora is structured into four subconditions defined by speaker and context: mother–home, mother–public, father–home, and father–public).  Monolingual baselines were formed by combining the relevant subsets (e.g., the English Baseline (Mom) consists of mother–home and mother–public data). Bilingual (random) datasets were constructed via probabilistic sampling at the dialogue level, while Bilingual (speaker) datasets align language with speaker identity, pairing mother–home and mother–public in one language (e.g., English) with father–home and father–public in the other (e.g., Spanish), and vice versa. Monolingual toplines and code-switching baselines were used as the full 100M corpora. All preprocessing methods used a fixed random seed of 42. 

For varying exposure experiments, we followed the method used to create the Bilingual (random) dataset by sampling at the dialogue level, but for each dialogue with both English and L2 versions, we selected the L2 instance with a specified probability (e.g. 0.9, 0.25, etc.) and the English instance otherwise. To construct the reduced data size experiments, we drew consistent random subsets of 20M and 10M word from the same underlying subcorpora using the same procedure. We preserved the relative structure of speaker–context conditions (e.g., mother vs.\ father, home vs.\ public), ensuring that the same 20M-word set of dialogue instances is used across language conditions.

Table~\ref{tab:dataset_stats} summarizes dialogue counts for each Spanish training condition, with counts approximately the same for Chinese and Dutch. The full Spanish datasets contain 799,885 dialogues in the 100M-word setting and 152,000 in the 20M-word setting, with corresponding baselines containing half as many dialogues and words. For each setting, we held out 5\% of dialogues for validation, using the same underlying dialogue instances across language variants to ensure comparability. Baseline validation sets were restricted to the corresponding 50\% subset of the full validation set.

\subsection{Training Details: Compute and Tokenization}\label{appendix:token}

Our experiments were conducted on NVIDIA A100 and H100 GPUs, using 8 GPUs per training run. Across approximately 400 training runs, each averaging 7–8 hours, we estimate a total computational budget of roughly 20,000–25,000 GPU-hours. This includes model pretraining across experimental conditions, model scales, and seeds. Models and tokenizers were trained with Transformers 4.29.0.dev0 and Tokenizers 0.13.3.

Tokenization is an important consideration for multilingual corpora: differences in word fragmentation and compression rates can cause comparable content to require more tokens in some languages than in others, affecting sequence lengths and the amount of text represented within a fixed token budget \citep{haslett-2025-tokenization, chang-etal-2024-multilinguality, wang2024tokenization, abagyan-etal-2026-one}. Accordingly, we trained a separate GPT-2 byte-level BPE tokenizer for each dataset condition to reflect its input distribution.

In early experiments, a vocabulary limit of 52k subword units was insufficient for the multilingual datasets, which exhibited greater lexical diversity than monolingual English (Appendix~\ref{appendix:datapreprocessing}). Accommodating this diversity within a fixed vocabulary can increase word fragmentation and impair downstream performance \citep{rust-etal-2021-good}. We therefore increased the maximum tokenizer vocabulary size to 80k tokens for all conditions \citep{chang-etal-2024-multilinguality, abagyan-etal-2026-one}. Actual vocabulary sizes varied: some conditions, such as the English baselines, did not reach this limit because they contained less data and only one language.\footnote{Larger tokenizer vocabularies also increase model parameter counts through the token embedding matrix.}

\subsection{Evaluation Details}\label{appendix:evaluationdatapreprocess}

We trained three models per condition using random seeds 42, 0, and 1. To limit the influence of words absent from the training data, we filtered the word similarity (WS) and Zorro evaluation sets using corpus vocabularies extracted with NLTK's \texttt{word\_tokenize} function. In particular, English evaluation items were restricted to words present in all three English baseline vocabularies (Mom, Dad, and random). We applied the same procedure to the 100M, 20M, and CHILDES settings. 

For WS, we additionally excluded words split into multiple tokens by the relevant tokenizers. This criterion differs from corpus vocabulary filtering: a word may occur in the training corpus yet still be fragmented, particularly if it is infrequent. Fragmentation requires averaging subword representations before computing cosine similarities and their Spearman correlation with human judgments, which can affect WS scores. Multilingual corpora have larger vocabularies and may therefore require larger tokenizer vocabularies to avoid excessive fragmentation. Consequently, lower WS scores largely would reflect tokenization effects rather than cross-lingual interference. This filtering complements the increased vocabulary limit described in Appendix~\ref{appendix:token} by directly excluding fragmented evaluation words.

The English WS benchmark used by \citet{zhuang2023visual} averages performance across five datasets: RG-65 \citep{rg-65}, WordSim-353 \citep{wordsim-353}, SimLex-999 \citep{hill-etal-2015-simlex}, SimVerb-3500 \citep{gerz-etal-2016-simverb}, and MEN (MTest-3000) \citep{bruni-etal-2012-distributional}. We additionally evaluated the English subset of MultiSimLex \citep{vulic-etal-2020-multi}. Weighting dataset scores by dataset size produced negligible differences relative to unweighted averaging; we therefore report unweighted means, consistent with \citet{zhuang2023visual}.

Our training dialogues contain speaker labels before utterances, whereas the standard evaluation items do not. We evaluated items without adding speaker labels by default and examined the effects of prepending labels in Appendix~\ref{appendix:impactspeakerlabels}.

As discussed in the main text (Figure \ref{fig:main_results}), Zorro scores can be sensitive to length normalization. \citet{liu-etal-2026-systematic} propose sub-linear length normalized log-probabilities (SLLN-LP) to mitigate biases arising from unequal sentence lengths in minimal pairs. We retain our original scoring method for consistency across evaluations. Normalization had similar, limited effects across trained conditions, with larger effects on randomly initialized and baseline models.

We used the LM Evaluation Harness implementations of the benchmarks cited in Section~\ref{sec:eval}, following established evaluation settings unless otherwise specified. Specifically, Zorro was evaluated using the BabyLM 2023 evaluation pipeline (\texttt{lm-evaluation-harness} v0.2.0, Transformers v4.26.1, PyTorch v2.2.0) in a zero-shot setting, with a null prompt template, no few-shot examples, a batch size of 1, and a random seed of 12. We report the unweighted mean accuracy across 13 Zorro categories. Word similarity was evaluated using Transformers v4.57.6, PyTorch v2.8.0, and SciPy v1.13.1. MultiBLiMP was evaluated using \texttt{minicons} v0.3.39, Transformers v5.16.1, and PyTorch v2.13.0.

\section{Additional Results}

\begin{table*}[t]
\centering
\small
\setlength{\tabcolsep}{3.2pt}

\resizebox{\textwidth}{!}{%
\begin{tabular}{lccc ccc cc}
\toprule
& \multicolumn{3}{c}{\textbf{Semantic Knowledge}}
& \multicolumn{3}{c}{\textbf{Grammatical Knowledge}}
& \multicolumn{2}{c}{\textbf{Perplexity}} \\
\cmidrule(lr){2-4}
\cmidrule(lr){5-7}
\cmidrule(lr){8-9}

\textbf{Model}
& \textbf{EN}
& \textbf{ES}
& \textbf{EN--ES}
& \textbf{EN}
& \textbf{ES}
& \textbf{H\textsubscript{EN,ES}}
& \textbf{EN}
& \textbf{ES} \\
\midrule

English Topline
& $0.58 \pm 0.01$
& $0.02 \pm 0.00$
& $0.08 \pm 0.00$
& $77.91 \pm 1.09$
& $62.50 \pm 3.24$
& $69.36 \pm 2.04$
& $2.43 \pm 0.00$
& $>10^{5}$ \\

Spanish Topline
& $0.07 \pm 0.02$
& $0.41 \pm 0.02$
& $0.05 \pm 0.00$
& $49.49 \pm 2.12$
& $78.04 \pm 0.48$
& $60.57 \pm 1.59$
& $>10^{3}$
& $2.64 \pm 0.01$ \\

\midrule

English Baseline (speaker)
& $0.54 \pm 0.01$
& $0.02 \pm 0.00$
& $0.08 \pm 0.00$
& $75.35 \pm 0.43$
& $59.86 \pm 0.90$
& $66.72 \pm 0.58$
& $4.60 \pm 0.17$
& $>10^{6}$ \\

English Baseline (random)
& $0.55 \pm 0.01$
& $0.01 \pm 0.00$
& $0.08 \pm 0.00$
& $75.41 \pm 1.00$
& $65.43 \pm 1.73$
& $70.07 \pm 1.08$
& $2.69 \pm 0.01$
& $>10^{6}$ \\

\midrule

Bilingual (speaker)
& $0.55 \pm 0.00$
& $0.38 \pm 0.01$
& $0.39 \pm 0.02$
& $75.71 \pm 0.89$
& $78.13 \pm 0.23$
& $76.90 \pm 0.47$
& $3.04 \pm 0.01$
& $3.37 \pm 0.02$ \\

Bilingual (random)
& $0.55 \pm 0.02$
& $0.36 \pm 0.03$
& $0.39 \pm 0.01$
& $74.51 \pm 2.20$
& $78.22 \pm 0.39$
& $76.32 \pm 1.17$
& $2.61 \pm 0.02$
& $2.82 \pm 0.02$ \\

\midrule
Code-switching (sentence)
& $0.56 \pm 0.01$
& $0.39 \pm 0.02$
& $0.40 \pm 0.03$
& $75.26 \pm 1.19$
& $78.92 \pm 0.37$
& $77.05 \pm 0.65$
& $5.26 \pm 0.56$
& $6.06 \pm 0.61$ \\

Code-switching (word)
& $0.55 \pm 0.01$
& $0.34 \pm 0.02$
& $0.38 \pm 0.01$
& $75.24 \pm 3.16$
& $76.24 \pm 0.50$
& $75.74 \pm 1.62$
& $3.19 \pm 0.05$
& $3.74 \pm 0.01$ \\

\bottomrule
\end{tabular}%
}

\caption{Evaluation results across semantic knowledge, grammatical knowledge, and perplexity. EN-ES denotes cross-lingual word similarity. H\textsubscript{EN,ES} denotes the harmonic mean of English and Spanish grammaticality. Values are mean $\pm$ standard deviation across seeds. This table is referred to in Section~\ref{sec:main_expt_results}.}
\label{tab:mainapp1}
\end{table*}

\begin{table*}[t]
\centering
\small
\setlength{\tabcolsep}{3.2pt}

\resizebox{\textwidth}{!}{%
\begin{tabular}{lccc ccc cc}
\toprule
& \multicolumn{3}{c}{\textbf{Semantic Knowledge}}
& \multicolumn{3}{c}{\textbf{Grammatical Knowledge}}
& \multicolumn{2}{c}{\textbf{Perplexity}} \\
\cmidrule(lr){2-4}
\cmidrule(lr){5-7}
\cmidrule(lr){8-9}

\textbf{Model}
& \textbf{EN}
& \textbf{ZH}
& \textbf{EN--ZH}
& \textbf{EN}
& \textbf{ZH}
& \textbf{H\textsubscript{EN,ZH}}
& \textbf{EN}
& \textbf{ZH} \\
\midrule

English Topline
& $0.59 \pm 0.02$
& $0.21 \pm 0.02$
& $0.04 \pm 0.01$
& $78.67 \pm 0.60$
& $49.06 \pm 4.04$
& $60.37 \pm 3.25$
& $2.43 \pm 0.01$
& $>10^{6}$ \\

Chinese Topline
& $0.12 \pm 0.02$
& $0.36 \pm 0.01$
& $-0.02 \pm 0.01$
& $51.97 \pm 4.05$
& $71.78 \pm 1.02$
& $60.22 \pm 2.76$
& $49.37 \pm 6.41$
& $5.25 \pm 0.01$ \\

\midrule

English Baseline (speaker)
& $0.56 \pm 0.00$
& $0.22 \pm 0.00$
& $0.06 \pm 0.01$
& $75.75 \pm 1.06$
& $48.51 \pm 1.47$
& $59.14 \pm 1.25$
& $3.14 \pm 0.04$
& $7.33 \pm 0.07$ \\

English Baseline (random)
& $0.58 \pm 0.01$
& $0.22 \pm 0.01$
& $0.05 \pm 0.01$
& $73.56 \pm 0.74$
& $50.74 \pm 1.53$
& $60.04 \pm 0.98$
& $2.67 \pm 0.01$
& $6.03 \pm 0.02$ \\

\midrule

Bilingual (speaker)
& $0.58 \pm 0.00$
& $0.33 \pm 0.01$
& $0.30 \pm 0.01$
& $75.54 \pm 1.83$
& $70.73 \pm 0.35$
& $73.05 \pm 0.83$
& $3.94 \pm 0.10$
& $9.56 \pm 0.43$ \\

Bilingual (random)
& $0.60 \pm 0.01$
& $0.36 \pm 0.02$
& $0.25 \pm 0.03$
& $74.42 \pm 0.64$
& $71.47 \pm 0.25$
& $72.92 \pm 0.39$
& $2.68 \pm 0.02$
& $6.05 \pm 0.06$ \\

\midrule

Code-switching (sentence)
& $0.59 \pm 0.01$
& $0.29 \pm 0.01$
& $0.34 \pm 0.03$
& $73.91 \pm 0.55$
& $72.76 \pm 0.95$
& $73.32 \pm 0.42$
& $5.72 \pm 0.41$
& $23.86 \pm 1.51$ \\

Code-switching (word)
& $0.58 \pm 0.00$
& $0.34 \pm 0.01$
& $0.35 \pm 0.02$
& $73.76 \pm 0.79$
& $71.33 \pm 0.54$
& $72.53 \pm 0.63$
& $3.03 \pm 0.01$
& $16.61 \pm 0.18$ \\

\bottomrule
\end{tabular}%
}

\caption{Evaluation results across semantic knowledge,
grammatical knowledge, and perplexity for English and Chinese.
EN--ZH denotes cross-lingual word similarity.
H\textsubscript{EN,ZH} denotes the harmonic mean of English
and Chinese grammaticality. Values are mean $\pm$ standard
deviation across three seeds. This table is referred to in Section~\ref{sec:main_expt_results}.}

\label{tab:mainapp2}
\end{table*}

\begin{table*}[t]
\centering
\small
\setlength{\tabcolsep}{3.2pt}

\resizebox{\textwidth}{!}{%
\begin{tabular}{lccc ccc cc}
\toprule
& \multicolumn{3}{c}{\textbf{Semantic Knowledge}}
& \multicolumn{3}{c}{\textbf{Grammatical Knowledge}}
& \multicolumn{2}{c}{\textbf{Perplexity}} \\
\cmidrule(lr){2-4}
\cmidrule(lr){5-7}
\cmidrule(lr){8-9}

\textbf{Model}
& \textbf{EN}
& \textbf{NL}
& \textbf{EN--NL}
& \textbf{EN}
& \textbf{NL}
& \textbf{H\textsubscript{EN,NL}}
& \textbf{EN}
& \textbf{NL} \\
\midrule

English Topline
& $0.57 \pm 0.01$
& $0.02 \pm 0.00$
& --
& $77.22 \pm 1.99$
& $54.40 \pm 0.97$
& $63.83 \pm 1.24$
& $2.42 \pm 0.00$
& $>10^{5}$ \\

Dutch Topline
& $0.09 \pm 0.01$
& $0.40 \pm 0.01$
& --
& $49.75 \pm 1.36$
& $80.22 \pm 0.44$
& $61.41 \pm 0.99$
& $>10^{3}$ 
& $2.75 \pm 0.01$ \\

\midrule

English Baseline (speaker)
& $0.55 \pm 0.00$
& $0.02 \pm 0.00$
& --
& $73.53 \pm 0.64$
& $53.22 \pm 0.85$
& $61.75 \pm 0.79$
& $4.71 \pm 0.08$
& $>10^{6}$ \\

English Baseline (random)
& $0.56 \pm 0.01$
& $0.02 \pm 0.00$
& --
& $74.92 \pm 1.74$
& $53.93 \pm 2.45$
& $62.71 \pm 2.26$
& $2.69 \pm 0.00$
& $>10^{6}$ \\

\midrule

Bilingual (speaker)
& $0.53 \pm 0.00$
& $0.34 \pm 0.02$
& --
& $76.45 \pm 1.41$
& $78.97 \pm 0.52$
& $77.69 \pm 0.82$
& $3.11 \pm 0.06$
& $3.46 \pm 0.02$ \\

Bilingual (random)
& $0.54 \pm 0.02$
& $0.36 \pm 0.01$
& --
& $74.67 \pm 1.91$
& $78.81 \pm 0.69$
& $76.67 \pm 1.08$
& $2.60 \pm 0.01$
& $2.96 \pm 0.00$ \\

\midrule

Code-switching (sentence)
& $0.54 \pm 0.01$
& $0.34 \pm 0.01$
& --
& $74.72 \pm 2.41$
& $79.35 \pm 0.48$
& $76.96 \pm 1.40$
& $4.83 \pm 0.12$
& $4.63 \pm 0.11$ \\

Code-switching (word)
& $0.51 \pm 0.01$
& $0.31 \pm 0.02$
& --
& $75.17 \pm 1.87$
& $79.82 \pm 0.29$
& $77.42 \pm 1.07$
& $3.39 \pm 0.01$
& $3.87 \pm 0.01$ \\

\bottomrule
\end{tabular}%
}

\caption{Evaluation results across semantic knowledge,
grammatical knowledge, and perplexity for English and Dutch.
H\textsubscript{EN,NL} denotes the harmonic mean of English
and Dutch grammaticality. Values are mean $\pm$ standard
deviation across three seeds. MultiSimLex for Dutch \citep{brans-bloem-2026-multi} does not provide a Dutch cross-lingual word similarity evaluation. This table is referred to in Section~\ref{sec:main_expt_results}.}

\label{tab:mainapp3}
\end{table*}

\subsection{Main Experiment}\label{app:main}
This section supplements Section~\ref{sec:main_expt_results} with tabular versions of Figure~\ref{fig:main_results} (Tables~\ref{tab:mainapp1}, \ref{tab:mainapp2}, and \ref{tab:mainapp3}). These tables additionally report cross-lingual word similarity and a joint multilingual grammaticality score, calculated as the harmonic mean of Zorro and MultiBLiMP/ZhoBLiMP performance. The final column of each table reports full perplexity values, complementing the abbreviated values presented in Table~\ref{tab:ppl} of the main text.

Notably, bilingual and code-switching models outperform monolingual toplines and English baselines on cross-lingual word similarity and joint multilingual grammaticality, indicating stronger cross-lingual semantic alignment and more balanced grammatical knowledge across languages.

\subsection{Reduced Model and Data Scale}\label{app:reducedscale}
\begin{figure*}[t]
\begin{center}
\includegraphics[width=\textwidth]{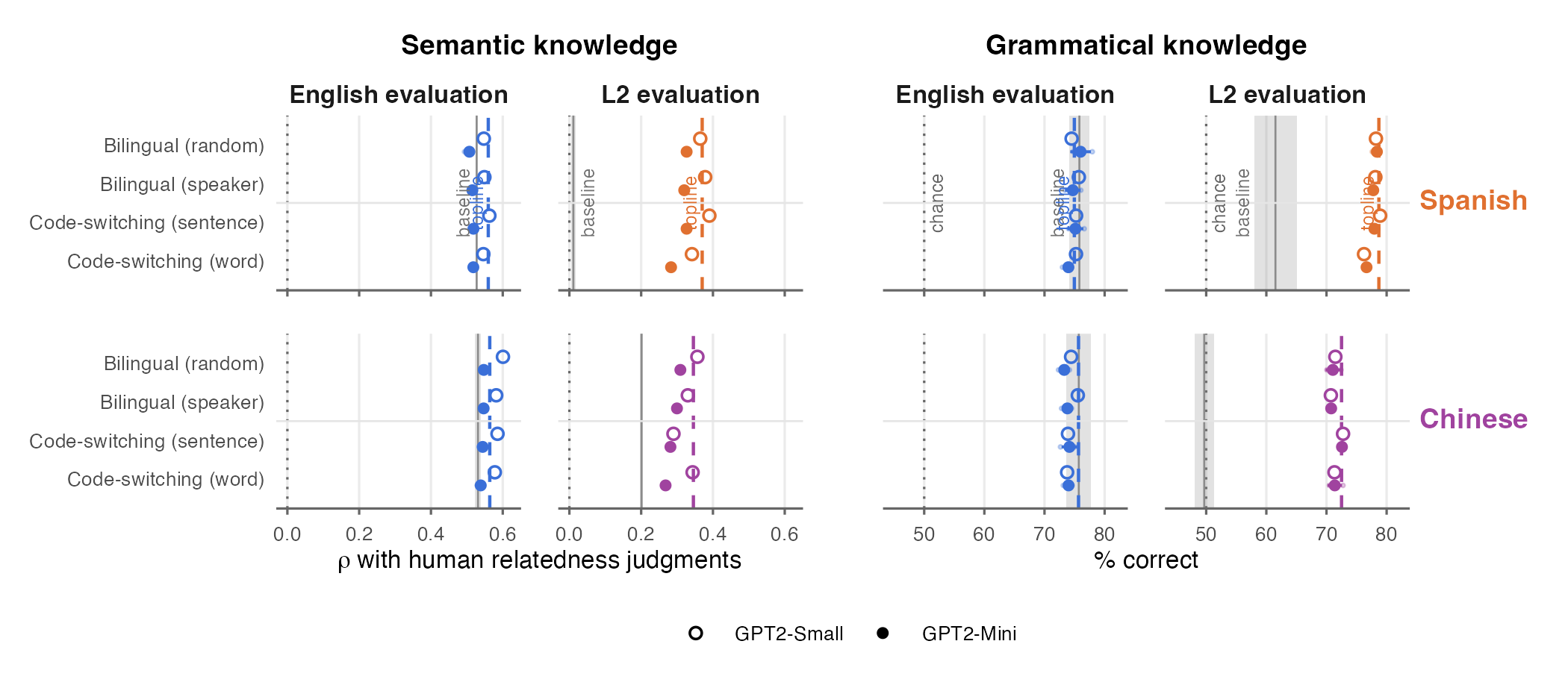}
\caption{Semantics and grammaticality evaluation results for GPT-2 Mini (39M) with GPT-2 Small (124M) reference markers. This figure is referred to in Section \ref{sec:reduced}.}\label{fig:gpt2mini}
\end{center}
\end{figure*}

\begin{figure*}[t]
\begin{center}
\includegraphics[width=\textwidth]{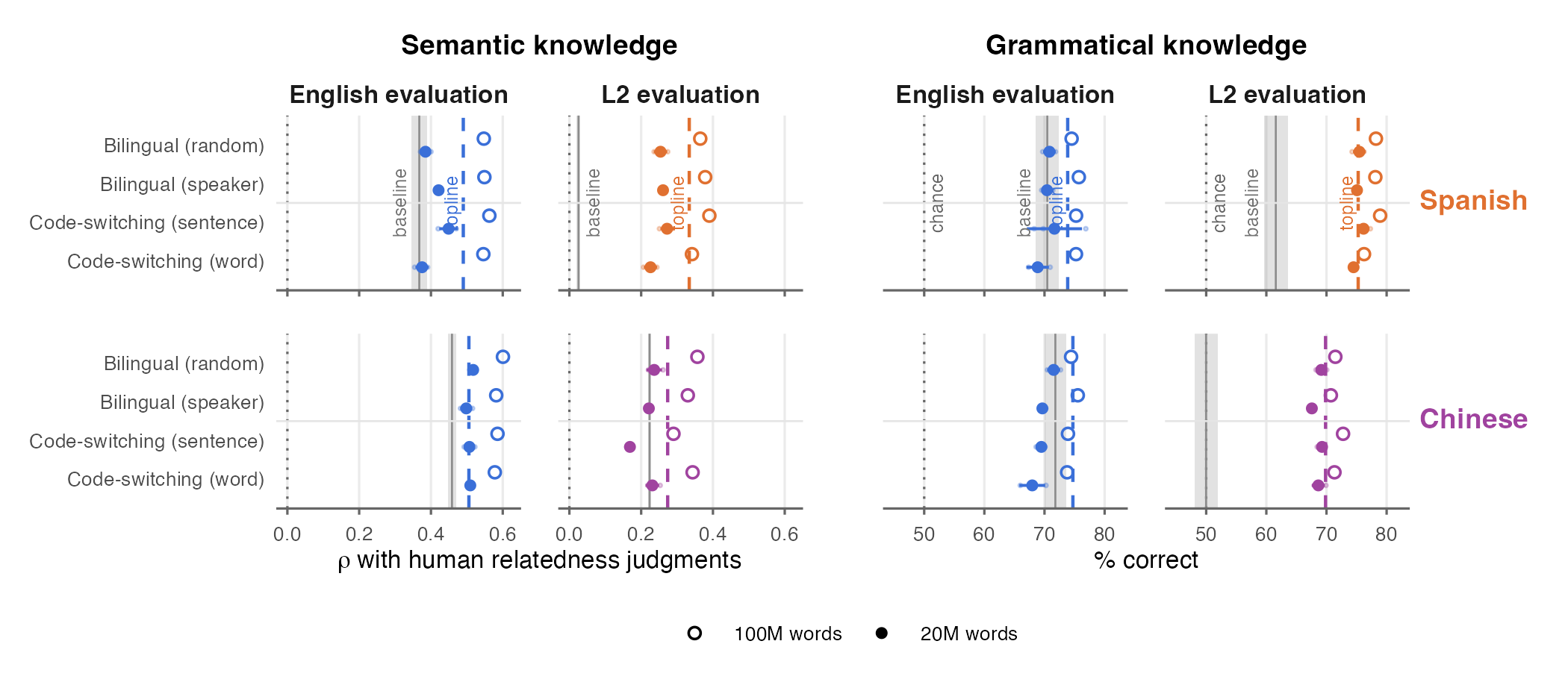}
\caption{Semantics and grammaticality evaluation results for GPT-2 Small (124M) trained on 20M words, with reference markers showing performance at 100M words. This figure is referred to in Section \ref{sec:reduced}.}\label{fig:20M}
\end{center}
\end{figure*}

\begin{figure*}[t]
\begin{center}
\includegraphics[width=\textwidth]{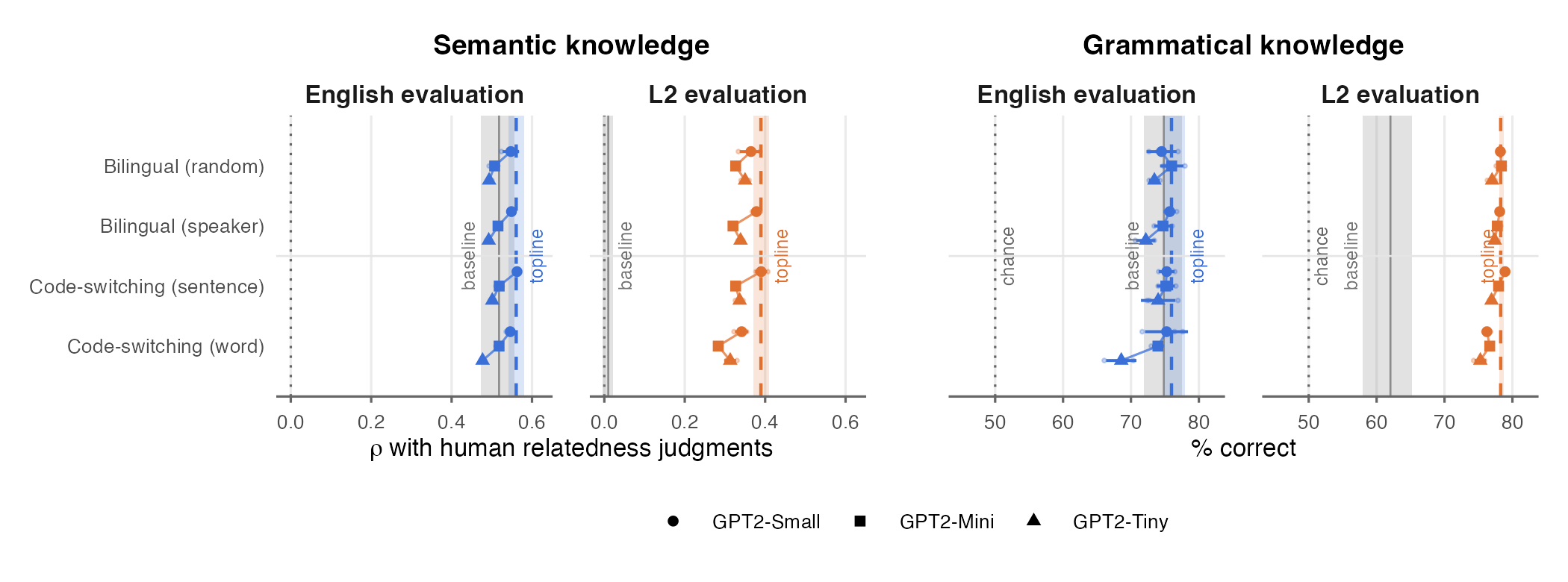}
\caption{Semantics and grammaticality evaluation results for GPT-2 across parameter scales in Spanish trained on 100M data points. This figure is referred to in Section \ref{sec:reduced}.}\label{fig:manyscales}
\end{center}
\end{figure*}

This section supplements Section~\ref{sec:reduced} with evaluation results for reduced model and data scale experiments in Spanish and Chinese (Figures~\ref{fig:gpt2mini} and~\ref{fig:20M}, respectively), as well as a comparison across three model scales in Spanish, including GPT-2 Tiny (Figure~\ref{fig:manyscales}).

For the model scaling experiments, we trained GPT-2 Mini (39M parameters) and GPT-2 Tiny (16M parameters) by reducing the number of transformer layers, attention heads, and hidden dimensions while keeping the training procedure consistent. For the data scaling experiments, we randomly subsampled 20M words from the original training corpora, using consistent samples across exposure conditions within each language pair. To preserve the exposure proportions of the main experiment, the English-only baseline was trained on 10M words rather than 20M, analogous to its 50M-word exposure in the original setting.

\subsection{The Role of Linguistic Structure}
This section supplements Section~\ref{sec:role} with evaluation results for the Spanish noise control experiments (Table~\ref{tab:es-controls}). Both the Shuffled Spanish and Random Characters conditions retain the 50M English words from the English Baseline (random) and replace the additional 50M Spanish words from the Bilingual (random) condition with noise. Specifically, Shuffled Spanish randomly permutes the order of Spanish words, while Random Characters shuffles individual characters. For the tokenizer-isolated controls, we use the tokenizer trained on the Bilingual (random) corpus rather than retraining it on the modified data, isolating the effects of noise exposure from changes in tokenization. These experiments use random seed 42.

\begin{table*}[t]
\centering
\small
\setlength{\tabcolsep}{4pt}

\resizebox{\textwidth}{!}{%
\begin{tabular}{lccc c c cc}
\toprule
& \multicolumn{3}{c}{\textbf{Semantic Knowledge}}
& \multicolumn{2}{c}{\textbf{Grammatical Knowledge}}
& \multicolumn{2}{c}{\textbf{Perplexity}} \\
\cmidrule(lr){2-4}
\cmidrule(lr){5-6}
\cmidrule(lr){7-8}

\textbf{Model / Condition}
& \textbf{EN}
& \textbf{ES}
& \textbf{EN--ES}
& \textbf{EN}
& \textbf{ES}
& \textbf{EN}
& \textbf{ES} \\
\midrule

English Baseline (random)
& $0.55 \pm 0.01$
& $0.01 \pm 0.00$
& $0.08 \pm 0.00$
& $75.41 \pm 1.00$
& $65.43 \pm 1.73$
& $2.69 \pm 0.01$
& $>10^{6}$ \\

Bilingual (random)
& $0.55 \pm 0.02$
& $0.36 \pm 0.03$
& $0.37 \pm 0.03$
& $74.51 \pm 2.20$
& $78.22 \pm 0.39$
& $2.61 \pm 0.02$
& $2.82 \pm 0.02$ \\

Shuffled Spanish
& $0.55 \pm 0.01$
& $-0.01 \pm 0.01$
& $0.11 \pm 0.01$
& $75.01 \pm 3.26$
& $66.93 \pm 0.54$
& $2.68 \pm 0.01$
& $62.06 \pm 0.12$ \\

Random Characters
& $0.44 \pm 0.01$
& $0.02 \pm 0.01$
& $0.05 \pm 0.01$
& $73.88 \pm 2.62$
& $65.64 \pm 0.53$
& $2.68 \pm 0.01$
& $>10^{3}$ \\

\midrule
\multicolumn{8}{l}{\textbf{Tokenizer-isolated controls}} \\
\midrule

Shuffled Spanish
& $0.55 \pm 0.01$
& $-0.01 \pm 0.01$
& $0.11 \pm 0.01$
& $74.99 \pm 3.17$
& $67.02 \pm 0.44$
& $2.67 \pm 0.01$
& $62.03 \pm 0.12$ \\

Random Characters
& $0.53 \pm 0.02$
& $0.00 \pm 0.01$
& $0.09 \pm 0.00$
& $73.69 \pm 2.94$
& $64.06 \pm 1.22$
& $2.67 \pm 0.01$
& $>10^{4}$ \\

\bottomrule
\end{tabular}%
}
\caption{Results of Spanish noise control experiments across semantic knowledge, grammatical knowledge, and perplexity evaluations. We compare the English-only baseline and bilingual condition against shuffled Spanish and random-character controls, including tokenizer-isolated variants that use a fixed tokenizer. Results are reported as mean $\pm$ standard deviation across three random seeds. This table is referred to in Section \ref{sec:role}.}\label{tab:es-controls}
\end{table*}

\subsection{Matching Tokenization Effects}
This section supplements Section~\ref{sec:match} with evaluation results from retraining key conditions using a shared tokenizer trained on the combined 100M-word English and 100M-word Spanish corpora (Table~\ref{tab:joint_tokenizer}). Using a shared tokenizer preserves the relative performance trends observed in the main experiments, with negligible changes in absolute performance (when comparing with Table \ref{tab:mainapp1}. The bilingual condition achieves comparable English semantic and grammatical performance to the English-only baseline (0.55 vs.\ 0.55 and 77.51 vs.\ 76.98, respectively), while substantially improving Spanish semantic knowledge (0.36 vs.\ $-0.01$), grammatical knowledge (78.47 vs.\ 67.10), and perplexity (2.82 vs.\ $>10^4$). Cross-lingual word similarity also improves from 0.10 to 0.40. These results suggest that the benefits of bilingual exposure are not driven by differences in tokenizer training across conditions, as the same trends persist when tokenization is held constant.

\begin{table}[t]
\centering
\small
\setlength{\tabcolsep}{3pt}

\resizebox{0.98\columnwidth}{!}{%
\begin{tabular}{lccc c c cc}
\toprule
& \multicolumn{3}{c}{\textbf{Semantics}}
& \multicolumn{2}{c}{\textbf{Grammaticality}}
& \multicolumn{2}{c}{\textbf{Perplexity}} \\
\cmidrule(lr){2-4}
\cmidrule(lr){5-6}
\cmidrule(lr){7-8}

\textbf{Condition}
& \textbf{EN}
& \textbf{ES}
& \textbf{EN--ES}
& \textbf{EN}
& \textbf{ES}
& \textbf{EN}
& \textbf{ES} \\
\midrule

English Topline
& 0.58
& 0.03
& 0.10
& 79.23
& 66.43
& 2.41
& $>10^{4}$ \\

Spanish Topline
& 0.02
& 0.36
& 0.09
& 48.75
& 80.76
& $>10^{4}$
& 2.62 \\

English Baseline (random)
& 0.55
& -0.01
& 0.10
& 76.98
& 67.10
& 2.67
& $>10^{4}$ \\

Bilingual (random)
& 0.55
& 0.36
& 0.40
& 77.51
& 78.47
& 2.60
& 2.82 \\

\bottomrule
\end{tabular}%
}

\caption{Evaluation results for Spanish when training with a joint tokenizer on all English Topline and Spanish Topline data (one run with seed 0), across semantic knowledge, grammatical knowledge, and perplexity metrics. This table is referred to in Section \ref{sec:match}.}
\label{tab:joint_tokenizer}
\end{table}

\subsection{Embedding Visualization Details}\label{appendix:viz}
Here, we provide additional details on token labeling for the embedding visualizations in Section~\ref{sec:modelinterp}. To generate the visualizations, we used Uniform Manifold Approximation and Projection (UMAP) with cosine distance, 30 nearest neighbors, and a minimum distance of 0.1. For each token in the model vocabulary, we computed the proportion of its occurrences in the English Topline and Spanish Topline corpora, both tokenized using the same model tokenizer. For computational efficiency, these proportions were estimated using a subset of 20{,}000 lines from each corpus rather than the full dataset of 799{,}885 lines. Tokens with at least 75\% of occurrences in one corpus were labeled English or Spanish, respectively; the remainder were labeled Shared. Tokens that did not appear in either corpus (e.g., rare subtokens or subtokens specific to certain multilingual models' tokenizers) were excluded. For computational efficiency and visual clarity, we restricted the visualizations to up to 20,000 tokens.

In Figure \ref{fig:embedding_grid}, overlapping regions tended to include common nouns (e.g., ``story'', ``poema''), potentially reflecting cross-lingual semantic alignment, and function or subword units (e.g., ``is'', ``mult''), whereas separated regions contain language-specific items, including less frequent vocabulary (e.g., ``bothering'', ``refrescante''). Meanwhile, the word-level code-switching condition (\ref{fig:embedding_grid}f) shows the global layout of the sentence-level condition, but the previously distinct English and Spanish clusters converge and overlap.

\subsection{Results with Speaker Labels}\label{appendix:impactspeakerlabels}
\cite{feng-etal-2024-child} demonstrate the importance of incorporating speaker labels during training. However, whether such labels are required at evaluation time to prevent domain shift remains unclear. To examine this, Table~\ref{tab:zorro_results} presents results from the original Zorro evaluation alongside an additional condition, Zorro (Mom), in which a fixed \texttt{**Mom**:} speaker label was prepended to every evaluation example across all Zorro datasets. We report Mom and Dad conditions separately without averaging them into by-speaker results.

Overall, introducing a fixed speaker label produces only minor shifts in performance. As expected, conditions aligned with the Mom speaker (e.g., English Baseline (Mom), Bilingual (Mom)) show slight improvements under Zorro (Mom), while Dad-aligned conditions tend to decrease. Importantly, besides these small variations, the relative ranking of conditions is preserved: the English Topline consistently achieves the highest performance, Spanish Topline shows the lowest performance, and the other conditions cluster within overlapping statistical intervals.

These findings suggest that speaker labels have a limited effect on downstream evaluation performance in this setting. As a result, we omit explicit speaker labels in subsequent Zorro experiments to avoid bias toward either parent speaker. Notably, this procedure is specific to the Zorro evaluation. For perplexity, speaker information is inherently present, as evaluation is performed directly on validation data from the English and Spanish toplines, which contain the same distribution of speaker labels as the training data. In contrast, word similarity evaluations directly compare two words, so they do not involve speaker labels.
\begin{table}[t]
\centering
\small
\resizebox{0.95\columnwidth}{!}{%
\begin{tabular}{lcc}
\toprule
\textbf{Name} & \textbf{Zorro} & \textbf{Zorro (Mom)} \\
\midrule
English Topline & 77.91 $\pm$ 1.09 & 79.06 $\pm$ 0.88 \\
Spanish Topline & 49.49 $\pm$ 2.12 & 47.89 $\pm$ 2.64 \\
\midrule
English Baseline (Mom) & 74.90 $\pm$ 1.89 & 76.07 $\pm$ 0.71 \\
English Baseline (Dad) & 75.96 $\pm$ 1.90 & 74.41 $\pm$ 0.54 \\
English Baseline (random) & 75.55 $\pm$ 1.23 & 76.79 $\pm$ 0.88 \\
\midrule
Bilingual (Mom) & 75.80 $\pm$ 0.78 & 77.06 $\pm$ 0.42 \\
Bilingual (Dad) & 75.62 $\pm$ 2.52 & 71.68 $\pm$ 0.81 \\
Bilingual (random) & 74.51 $\pm$ 2.20 & 75.77 $\pm$ 1.94 \\
\midrule
Code-switching (sentence) & 75.26 $\pm$ 1.19 & 76.01 $\pm$ 1.81 \\
Code-switching (word) & 75.24 $\pm$ 3.16 & 76.24 $\pm$ 3.26 \\
\bottomrule
\end{tabular}%
}
\caption{Results evaluating the effect of including a speaker label during Zorro evaluation across all language conditions. For multilingual conditions, the label in parentheses (e.g., Bilingual (mom)) indicates that Mom is assigned to the English-speaking role. Results are reported as mean $\pm$ SD across runs.}
\label{tab:zorro_results}
\end{table}

\section{LLM Usage Disclosure}

We used LLMs (e.g., GPT-5) to assist with parts of the coding process and limited aspects of paper preparation, such as checking grammar. All outputs were carefully reviewed to ensure accuracy and appropriateness.

\end{document}